\PassOptionsToPackage{section}{placeins}
\documentclass{wechat}
\usepackage{amsmath,amssymb,booktabs,tabularx,array,graphicx}
\usepackage{enumitem}
\usepackage{url}
\expandafter\def\expandafter\UrlBreaks\expandafter{\UrlBreaks\do\-}
\usepackage{longtable,colortbl}
\providecommand{\doi}[1]{\href{https://doi.org/#1}{\nolinkurl{doi:#1}}}
\newcommand{\model}{WeMM-Ecom}

\graphicspath{{figs/}}
\DeclareCaptionFont{overview}{\fontsize{8.6}{10.3}\selectfont}
\setlist{nosep,leftmargin=*}
\newenvironment{papertablenotes}{%
  \par\smallskip\noindent\begin{minipage}{\linewidth}%
  \footnotesize\raggedright
  \begin{itemize}[label={},labelsep=0pt,labelwidth=0pt,leftmargin=0pt,nosep]%
}{%
  \end{itemize}\end{minipage}%
}
\newcolumntype{Y}{>{\centering\arraybackslash}X}

\newcommand{\linkicon}[1]{\raisebox{-0.2em}{\includegraphics[height=1em]{#1}}}
\newcommand{\hficon}{\linkicon{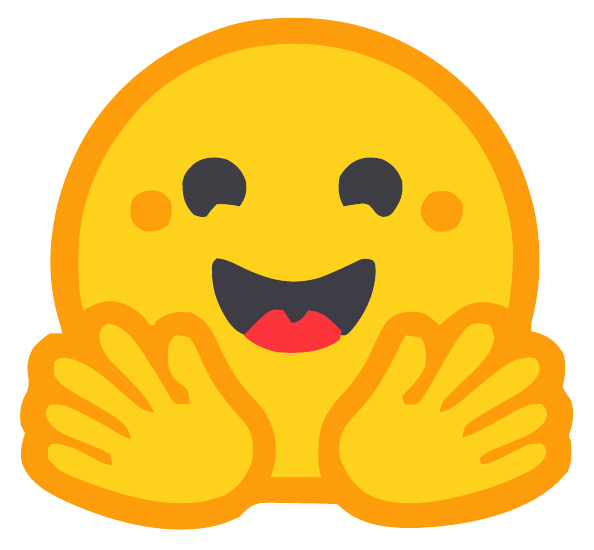}}
\newcommand{\ghicon}{\linkicon{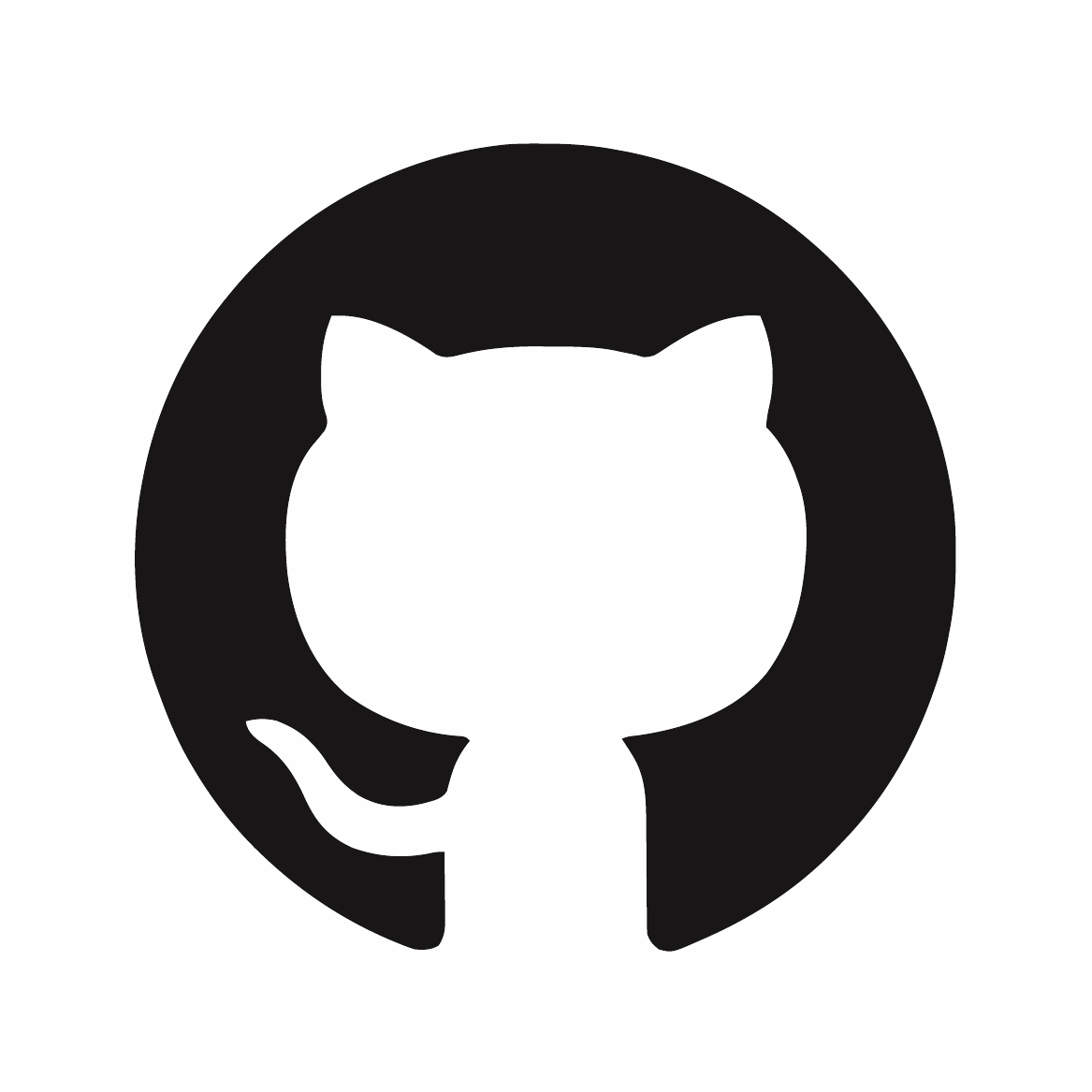}}
\newcommand{\hflink}{https://huggingface.co/collections/tencent/wemm-ecom}
\newcommand{\ghlink}{https://github.com/Tencent/WeMM-Ecom}
\newcommand{\resourcelinks}{%
  \begingroup
  \small\raggedright
  \setlength{\parindent}{0pt}%
  \setlength{\parskip}{0pt}%
  \href{\hflink}{\makebox[1.6em][l]{\hficon}\texttt{\hflink}}\par
  \href{\ghlink}{\makebox[1.6em][l]{\ghicon}\texttt{\ghlink}}\par
  \endgroup
}

\title{WeMM-Ecom: Advancing Multimodal Content Understanding for WeChat E-Commerce}
\author{%
\fontsize{11.2pt}{13.2pt}\selectfont
\centerline{%
   Binhui Xie$^{*}$, Jinmin Li$^{*}$, Haomiao Tang$^{*}$, Guangting Wang$^{*\dagger}$}
\centerline{
    Song Jin, Jianwei Zhu, Fengyun Rao$^{\ddagger}$, Jing LYU
}
\vspace{6pt}
\centerline{%
   WeChat Vision, Tencent Inc.
}
\vspace{-6pt}
}
\abstract{
We present \model{}, a family of multimodal large language models that deliver frontier e-commerce capabilities while achieving state-of-the-art general-purpose performance at their respective scales. The family spans five sizes, from a 0.8B dense model to a 35B mixture-of-experts model with 3B active parameters. Our training pipeline begins with two-stage continued pre-training that progressively raises the proportion of e-commerce data within a diverse multimodal data mixture, followed by hybrid-thinking post-training that combines on-policy distillation and reinforcement learning to support both non-thinking and thinking modes. For evaluation, we construct a comprehensive suite of 28 e-commerce benchmarks (23 in-house and five public) covering eight business scenarios and complement it with 64 public general-purpose benchmarks. On the e-commerce suite, WeMM-Ecom-35B-A3B surpasses substantially larger open-source flagship models, including Qwen3.8-Flash-Next, HY4, and DeepSeek-V4.1-Flash-1017. On the general-purpose suite, every \model{} model surpasses its similarly sized Qwen3.5 counterpart, with the lightweight variants (e.g., WeMM-Ecom-2B) showing clear advantages over leading lightweight models such as Qwen3.5-2B, MiniCPM-V-4.7-1B, and LFM3-VL-2B. Beyond benchmarks, \model{} now powers hundreds of production use cases in WeChat, including catalog management, multimodal product search, and livestream commerce, serving billion-scale requests per day. We publicly release all model weights and detail our data curation, training recipe, and benchmark construction to facilitate future work on domain-specialized multimodal models.

\par\vspace{5pt}
\resourcelinks
\vspace{-3.5pt}
}
\DeclareCaptionLabelFormat{casepanel}{Figure~\thefigure(#2)}
\begin{document}
\maketitle
\enlargethispage{11pt}

\begin{minipage}{\linewidth}
\centering
\makebox[\linewidth][c]{\includegraphics[width=0.94\linewidth]{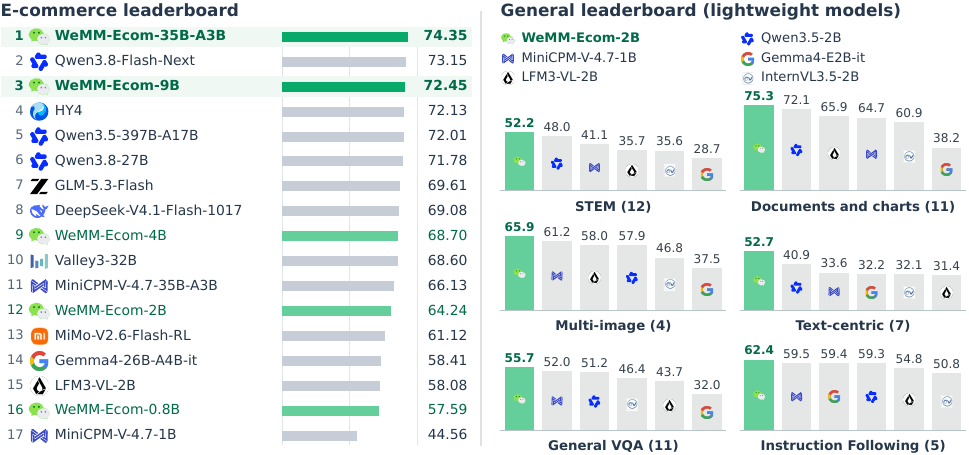}}
\captionsetup{font=overview,skip=2pt}
\captionof{figure}{\textbf{Comparing WeMM-Ecom against SoTA multimodal models.} On the e-commerce suite, WeMM-Ecom-35B-A3B surpasses models $5\times$--$22\times$ larger (e.g., \mbox{Qwen3.8-Flash-Next-180B} and \mbox{HY4-770B}) (left). On the general-purpose suite, WeMM-Ecom-2B scores highest among lightweight models across six domains in non-thinking mode (right).}
\label{fig:headline_results}
\end{minipage}

\begingroup
\renewcommand{\thefootnote}{}
\footnotetext{\scriptsize
$^{*}$Core contributors.
\quad
$^{\dagger}$Project leader.
\quad
$^{\ddagger}$Corresponding author: \href{mailto:fengyunrao@tencent.com}{\nolinkurl{fengyunrao@tencent.com}}
}
\endgroup

\section{Introduction}
\label{sec:intro}

Multimodal large language models (MLLMs) have become capable general-purpose models for understanding images, documents, and video~\citep{hurst2024gpt,team2025kimi,qwenteam2026qwen35,huang2026step3,hong2025glm,wang2025internvl3,team2026gemma}. In specialized domains, however, general capability alone falls short, as models must apply fine-grained domain knowledge to what they see, whether identifying a clinical finding in a medical scan~\citep{sellergren2026open} or extracting the correct figure from a financial chart~\citep{luo2025finmme}.

E-commerce is an especially demanding domain for MLLMs. A product can be presented through dozens of product views, long-form detail images, and livestream frames, alongside titles and structured attribute tables. Critical evidence, such as selling points, nutrition facts, dimension annotations, and certification marks, often appears only inside images. Models must further link this visual text to the specific SKU variant in question and draw on catalog knowledge that is never explicitly stated. As e-commerce models move from text records~\citep{ma2023ecomgpt,li2024ecomgpt,peng2024ecellm} to multimodal understanding~\citep{wu2025valley2,chen2026valley3}, two core challenges stand out: (1) models must accurately perceive dense, interleaved visual evidence and produce outputs that satisfy strict entity-level and business-rule constraints; and (2) domain specialization must not erode general visual perception and reasoning, especially for lightweight models deployed under tight latency and memory budgets~\citep{yu2026minicpm,liquidAI2026LFM3}.

To address these challenges, we present \textbf{\model}, a family of multimodal foundation models spanning four dense sizes (0.8B, 2B, 4B, and 9B) and a 35B mixture-of-experts (MoE) model with 3B active parameters (35B-A3B), initialized from Qwen3.5~\citep{qwenteam2026qwen35}. We curate over 40 e-commerce tasks from product images, livestream frames, and business workflows. Because these tasks directly exercise visual grounding, OCR, and multi-image reasoning, they also strengthen general perception rather than competing with it. We pair them with broad general supervision in a two-stage continued pre-training (CPT) curriculum with an increased e-commerce share in the second stage (Figure~\ref{fig:cpt_data_mixture}). Post-training then applies hybrid-thinking on-policy distillation (OPD), a single parameter-space merge with the post-trained Qwen3.5 weights that recovers general instruction-following and reasoning abilities, and group relative policy optimization (GRPO), enabling each model to support both thinking and non-thinking modes.

We evaluate \model\ on a comprehensive 28-benchmark e-commerce suite across eight scenarios (207,117 instances) and 64 general benchmarks. Across all five sizes and both modes, \model\ consistently outperforms size-matched Qwen3.5 on e-commerce. WeMM-Ecom-35B-A3B ranks first among evaluated flagships (Figure~\ref{fig:headline_results}), surpassing Qwen3.8-Flash-Next, HY4, and DeepSeek-V4.1-Flash-1017. In non-thinking mode, our 0.8B and 2B models each beat a larger Qwen3.5 model on e-commerce while showing clear advantages on general-purpose capabilities. WeMM-Ecom-2B also surpasses MiniCPM-V-4.7-1B~\citep{yu2026minicpm} and LFM3-VL-2B~\citep{liquidAI2026LFM3} on both e-commerce and general suites in both thinking and non-thinking modes.

Our main contributions are as follows:
\begin{enumerate}
\item \textbf{We open-source \model{}}, a family of five multimodal models ranging from a 0.8B dense model to a 35B-A3B MoE model, each supporting both thinking and non-thinking modes. \model{} delivers leading e-commerce performance at every scale while matching or surpassing its Qwen3.5 counterpart on general benchmarks, and has been validated at scale in production.
\item \textbf{We present a detailed training and data recipe for \model{}}. We describe how more than 40 e-commerce tasks are constructed from product images, livestream frames, and business records; how they are mixed with general data and scheduled across two CPT stages; and how hybrid-thinking OPD, a one-time parameter-space merge, and GRPO are combined to produce the final models.
\item \textbf{We introduce the key design principles of our e-commerce evaluation framework}. We show how real-world business workflows are translated into eight evaluation scenarios with explicit inputs, decision targets, and scoring rules, resulting in a suite of 23 in-house and five public benchmarks with 207,117 instances.
\end{enumerate}
\captionsetup[figure]{font=normalsize}
\captionsetup[subfigure]{font=normalsize}
\section{Training}
\label{sec:methods}
\label{sec:method}
\begin{figure}[tbp]
\centering
\includegraphics[width=\linewidth]{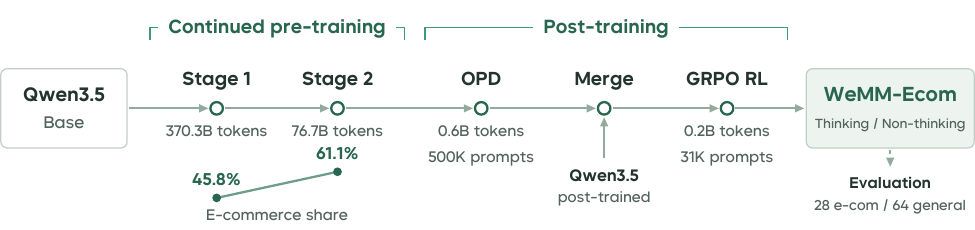}
\caption{\textbf{Training and evaluation pipeline.} The e-commerce share increases in the second continued pre-training stage. Post-training combines on-policy distillation (OPD), checkpoint merging, and group relative policy optimization (GRPO). The resulting checkpoints support thinking and non-thinking modes and are evaluated on e-commerce and general benchmarks.}
\label{fig:training_pipeline}
\end{figure}

We initialize each \model\ model from the corresponding Qwen3.5 Base checkpoint~\citep{qwenteam2026qwen35} and perform full-parameter optimization across all training stages using the WeChat-YATT framework~\citep{wu2025wechat}. Our pipeline first conducts CPT to instill fine-grained knowledge of products, variants, and business rules while preserving broad multimodal capabilities. Subsequently, post-training combines OPD for token-level teacher guidance, a single parameter-space merge with the post-trained Qwen3.5 weights, and reinforcement learning (RL) driven by task-specific verifiable rewards. All five model scales support thinking and non-thinking modes in a unified chat interface. Figure~\ref{fig:training_pipeline} summarizes the training and evaluation pipeline.

\subsection{Continued Pre-Training for E-Commerce}
\label{sec:cpt}

\begin{figure}[!htbp]
\centering
\includegraphics[width=\linewidth]{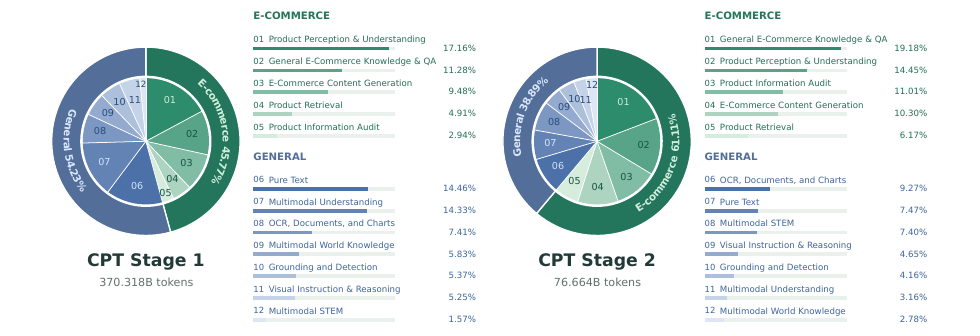}
\caption{\textbf{Token composition of the two CPT stages.} Inner sectors and adjacent bars show shares across five e-commerce and seven general domains; outer rings show proportion of general and e-commerce data.  E-commerce accounts for 45.8\% of 370.32B tokens in Stage~1 and 61.1\% of 76.66B tokens in Stage~2. Domains are ranked independently within each group and stage, thus number indices are stage- and group-specific.}
\label{fig:cpt_data_mixture}
\end{figure}

\begin{table}[!htbp]
\centering
\small
\setlength{\tabcolsep}{6pt}
\renewcommand{\arraystretch}{1.15}
\caption{\textbf{CPT domains.} Each record is counted once, in the domain of
its primary supervision objective rather than by source name or number of
images. Records from e-commerce business sources remain in the e-commerce pool
regardless of modality, and general records without visual input form Pure
Text. Per-stage proportions are shown in Figure~\ref{fig:cpt_data_mixture}.}
\label{tab:cpt_domains}
\begin{tabular}{@{}>{\raggedright\arraybackslash}p{0.27\linewidth}>{\raggedright\arraybackslash}p{0.69\linewidth}@{}}
\toprule
Domain & Scope \\
\midrule
\multicolumn{2}{@{}l}{\textit{E-commerce}} \\
Product Perception \& Understanding &
  Category--property--value (CPV) prediction, including category prediction
  with reasoning, attribute extraction, and SKU specification understanding,
  together with localization and detection of target products in listing
  images, multi-image listings, and livestream frames. \\
Product Information Audit &
  Rule-conditioned review of product listings: consistency across images,
  titles, categories, and attributes; category mismatch; product-image
  quality; inappropriate content; and quality checks of short titles. \\
Product Retrieval &
  Same-product matching and retrieval at SPU or SKU granularity, including
  image-to-image and image-to-product search, text-based SKU matching, and
  similarity judgments between visual entities or product regions. \\
General E-Commerce Knowledge \& QA &
  Question answering over product images and listing text, OCR of product and
  storefront images, shopping and price reasoning, and product, brand, and
  entity knowledge, including entity recognition and product-word prediction. \\
E-Commerce Content Generation &
  Product-centric text generation: per-view and overall descriptions of
  product images, multi-image listings, and livestream frames; short product
  titles; and product copy and product-sharing posts. \\
\midrule
\multicolumn{2}{@{}l}{\textit{General}} \\
Multimodal Understanding &
  Foundational understanding of single images, multiple images, and videos:
  dense image captions, per-image and overall multi-image descriptions, video
  event understanding and QA, and video temporal grounding. \\
OCR, Documents, and Charts &
  Text-rich inputs: text transcription, including multilingual OCR; document
  and table parsing; document QA; chart reading; and structured extraction of
  visible text. \\
Grounding and Detection &
  Image-space localization with coordinate targets: object detection,
  referring-expression grounding, GUI element grounding, and rejection of
  queries whose target is absent. \\
Visual Instruction \& Reasoning &
  Visual tasks beyond description: VQA and cross-image comparison, counting
  and relational questions without coordinate targets, image-quality
  assessment, inferring edit instructions from source--edited image pairs, and
  rule-conditioned moderation of general (non-listing) content. \\
Multimodal STEM &
  Mathematics, science, and technical problems that require visual context
  (diagrams, charts, documents, educational material, or video), with worked
  solutions; includes chart-to-code tasks. \\
Multimodal World Knowledge &
  Linking visual entities and scenes to external facts: encyclopedia- and
  entity-grounded image descriptions, knowledge-intensive VQA, and
  specialized-domain tasks such as medical imaging. \\
Pure Text &
  General text-only records: mathematical and scientific reasoning,
  instruction-following and multilingual dialogue, encyclopedic and
  reading-comprehension text, and curated web text. \\
\bottomrule
\end{tabular}
\end{table}

CPT~\citep{liu2025midtraining} adapts the backbone to e-commerce while preserving its general multimodal capabilities. We construct CPT data from product images, livestream frames, business records, and general multimodal and text-only corpora, and organize it into five e-commerce and seven general domains (Figure~\ref{fig:cpt_data_mixture}; Table~\ref{tab:cpt_domains}). We retain both pools throughout CPT because e-commerce tasks routinely depend on general skills: reading a package label and assigning it to the requested product, for example, draws on general OCR and grounding as well as knowledge of product variants. Records with an e-commerce business context, including text-only records, are assigned to the corresponding e-commerce domain.
Training proceeds in two stages; the second raises the e-commerce share and shifts emphasis from perception toward auditing and reasoning (Section~\ref{sec:cpt_recipe}).

\subsubsection{E-Commerce Data}
\label{sec:ecom_data}
General multimodal corpora teach a model to recognize a bag of coffee beans;
e-commerce applications must also determine which variant it is---the 250\,g
or the 500\,g SKU, a single bag or a two-pack---whether its listing complies
with platform rules, and how to present it to shoppers. We therefore design
e-commerce supervision around three properties that general corpora rarely
provide: \textit{variant-level granularity}, \textit{traceable evidence}, and \textit{rule-conditioned
outputs}. The five domains in Table~\ref{tab:cpt_domains} instantiate these
properties in different combinations.

\paragraph{Variant-level granularity.}
Product Perception \& Understanding trains the model to resolve products at the
level of categories, attributes, and SKUs. Category prediction pairs product
images with taxonomy labels and reasoning traces, and attribute tasks extract
visible text, properties, and regions from gallery images, detail strips, and
livestream frames. SKU attribute examples specify a variant and an attribute,
supply the relevant views and fields, and use the recorded value as the target;
a net-weight question, for instance, must distinguish the 250\,g SKU from the
500\,g SKU. Product Retrieval applies the same granularity to matching. Each
example pairs two listings, SKU records, or product regions with their
available images and specifications, and defines equivalence at the requested
level: a single 500\,g bag and a two-pack of 250\,g bags may depict the same
product and contain the same total quantity, yet their pack configurations make
them distinct SKUs. Region-level and cross-image examples extend matching to
comparisons between specified visual entities.

\paragraph{Traceable evidence.}
We tie targets to the evidence that supports them so that outputs can be
checked against the source listing. SKU attribute examples retain the index of
the source image that contains the recorded value. Bounding-box annotations
localize products, brand logos, certification marks, specification tables, and
dimension annotations, and explicitly mark targets that are absent from the
image. In General E-Commerce Knowledge \& QA, listing images, descriptions,
visible text, and structured fields are paired with questions on product
knowledge, commerce OCR, and reasoning, with reference answers supported by the
supplied sources. Product descriptions, entity annotations, and articles
further connect product appearance with names, brands, uses, and business
knowledge.

\paragraph{Rule-conditioned outputs.}
Many e-commerce decisions are defined by platform rules rather than by visual
content alone. Product Information Audit pairs listing titles, gallery images,
and attribute fields with platform governance rules and annotated compliance
decisions. Its tasks include identifying restricted goods, detecting
image--title or category inconsistencies, reviewing product-image quality, and
screening content for policy violations; each target specifies the applicable
rule and the decision, together with visual or textual evidence for tasks that
require an explanation. The same domain also covers quality review of short
product titles, where the target is a set of pass/fail judgments rather than a
rewritten title. E-Commerce Content Generation applies rules to generation
rather than judgment: short-title examples map structured product
specifications to concise, formatted titles that follow platform style
guidelines.

\subsubsection{General-Purpose Data}
\label{sec:data}
The seven general domains preserve the perception, language, knowledge, and
reasoning capabilities that e-commerce tasks build on and that the model needs
beyond them. They draw on curated sources and task mixtures, including all 162
nonempty sources in OpenBee~\citep{zhang2026bee}. We summarize the main sources
below and describe shared construction procedures in
Section~\ref{sec:data_quality}.

\paragraph{Perception and reading.}
Multimodal Understanding pairs images with Chinese or English captions and adds
video data for event understanding and temporally grounded tasks, including
sports event narration and general video question answering. OCR, Documents,
and Charts pairs natural images, documents, charts, and tables with
transcriptions, reading questions, or structured extraction targets, covering
both the recognition of visible text and its use in document and chart
understanding; sources include ChartNet~\citep{kondic2026chartnet},
multilingual OCR collections, web articles, and table annotations.
Grounding and Detection converts object and region annotations from
Objects365~\citep{shao2019objects365}, SA-1B~\citep{kirillov2023segment}, and
LocateAnything~\citep{nvidia2026locateanything} into localization and detection
targets conditioned on object names or referring descriptions, while GUI
examples associate interface instructions with the locations of the
corresponding elements.

\paragraph{Reasoning and knowledge.}
Visual Instruction \& Reasoning combines single-image QA, multi-image
comparison, visual quality assessment, and instruction-driven tasks. Each
example pairs the available views with a question or constrained instruction
and a reference answer, comparison judgment, or decision; multi-image questions
require integrating evidence across views, and edit-instruction examples ask
the model to infer, from a source image and its edited version, the
instruction that produced the change. Risk assessment and content moderation
tasks apply written platform policies to general content such as short videos;
unlike Product Information Audit, they do not review product listings.
Multimodal STEM pairs diagram- and video-based problems in mathematics,
science, and technical reasoning with worked solutions, including material from
Innovator-VL~\citep{wen2026innovator}, and adds chart and document problems
from educational corpora across biology, chemistry, mathematics, and K-12
science. Multimodal World Knowledge associates images with encyclopedic and
entity-specific descriptions from online wikis and entity recaption sources, and pairs
visual evidence with specialized questions whose answers draw on knowledge
beyond the visible appearance.

\paragraph{Text-only data.}
Pure Text retains general records without visual inputs as language and
instruction supervision. Mathematical and scientific reasoning dominate this
mixture, drawing on Nemotron-Math~\citep{du2025nemotron} and Chinese STEM
sources, while instruction-following and multilingual dialogues, Wikipedia
text, reading-comprehension QA, and curated web text supply general language,
factual knowledge, and instruction-following supervision.

\subsubsection{Data Construction and Quality Control}
\label{sec:data_quality}
Several construction and filtering procedures are shared across domains.

\paragraph{Spatial targets and absent objects.}
Conversion of box annotations in both pools makes coordinate normalization
explicit, preserves the source category schemas, and keeps each target linked
to its annotated visual entity. Both pools also encode absence explicitly:
e-commerce annotations mark targets missing from the image, and general
grounding data includes rejection examples whose queries name objects absent
from the image, teaching the model to decline rather than hallucinate a box.

\paragraph{Recaptioning.}
We cluster images with WeMM-Embedding~\citep{zhou2026wemm}, sample across
clusters for recaptioning, and deduplicate the resulting captions. For
text-rich images, PaddleOCR~\citep{cui2025paddleocr} transcriptions provide
textual cues for caption generation.

\paragraph{Synthesized reasoning.}
For synthesized STEM solutions, Qwen3.5-122B-A10B generates the reasoning and
Qwen3.5-35B-A3B checks answer agreement; malformed or truncated solutions are
discarded.

\paragraph{Web article mining.}
Two classifiers identify product-recommendation articles and select
comparative reviews or detailed accounts of product use. Qwen3.5-122B-A10B then
filters promotional or irrelevant material and extracts product and brand
facts, which we deduplicate with MinHash.

\paragraph{Source balancing.}
When aggregating multi-source collections, we assign a token budget to each
source group and, where a group exceeds its budget, downsample its constituent
sources in proportion to their token counts, so that every source remains
represented.

\subsubsection{CPT Recipe}
\label{sec:cpt_recipe}

\paragraph{Progressive data mixture.}
We adopt a two-stage curriculum that increases domain specialization while
maintaining broad general supervision (Figure~\ref{fig:cpt_data_mixture}). Stage~1 establishes multimodal and
product representations with a 45.8\% e-commerce share, and Stage~2 raises
the share to 61.1\%. Within the e-commerce pool (percentages are shares within
each pool), supervision shifts from basic perception toward compliance auditing
and knowledge-intensive QA: Product Information Audit rises from 6.4\% to
18.0\%, General E-Commerce Knowledge \& QA from 24.7\% to 31.4\%, and Product
Perception \& Understanding falls from 37.5\% to 23.6\%. Within the general
pool, supervision shifts from captioning and text-only data toward dense text
reading and visually grounded reasoning: OCR, Documents, and Charts rises from
13.7\% to 23.8\% and Multimodal STEM from 2.9\% to 19.0\%, while Multimodal
Understanding falls from 26.4\% to 8.1\% and Pure Text from 26.7\% to 19.2\%.
Throughout CPT, we track per-domain validation losses to guide mixture adjustments. 

\paragraph{Training schedule and objective.}
All model sizes follow the same two-stage CPT schedule: Stage~1 trains on 370.3B tokens at a 16K sequence length, and Stage~2 extends the context window to 32K for 76.7B tokens. For a training batch $\mathcal B$, let $v_i$ denote the visual inputs of example $i$, $z_i$ its token sequence, and $\mathcal M_i$ its supervised response token positions with $N_i=|\mathcal M_i|>0$. To prevent long responses from dominating gradient updates while still accounting for sequence length, we employ square averaging~\citep{chen2024expanding,wang2025internvl3}, weighting each supervised token by $N_i^{-1/2}$:
\begin{equation}
 \mathcal L_{\mathrm{CPT}}(\theta;\mathcal B)
 =-\frac{\sum_{i\in\mathcal B}N_i^{-1/2}\sum_{t\in\mathcal M_i}
 \log\pi_\theta(z_{i,t}\mid v_i,z_{i,<t})}
 {\sum_{i\in\mathcal B}\sqrt{N_i}}.
 \label{eq:cpt}
\end{equation}
We train with dynamic micro-batching, visual-encoder parallelism, and sequence packing up to each stage's context limit, preserving per-example loss masks and image--text alignments.

\subsection{Post-Training}
Following CPT, our post-training pipeline combines OPD, parameter-space model merging, and GRPO-based reinforcement learning. Both post-training stages optimize a single set of weights to support both thinking and non-thinking modes~\citep{yu2026minicpm}.

\subsubsection{Post-Training Data}
The post-training corpus spans general reasoning, visual perception, and instruction following alongside core e-commerce workflows such as variant matching and product detection. All model sizes share a curated pool of 499,998 OPD prompts---including specialist data for open-vocabulary detection, complex counting, and long-document QA---and 30,976 prompts for RL. We select OPD prompts via offline teacher--student comparisons, retaining instances where the teacher achieves a higher mean quality score than the student. For RL, we sample multiple rollouts from the corresponding post-OPD merged checkpoint and retain prompts with positive reward variance while filtering out already saturated instances (Appendix~\ref{app:post_training_screening}).

\subsubsection{Hybrid Thinking}
To support both deliberative chain-of-thought reasoning (thinking mode) and low-latency direct responses (non-thinking mode) within a single checkpoint, both OPD and RL assign training prompts to thinking and non-thinking modes at a $4{:}1$ ratio (Appendix~\ref{app:mode_assignment}). Because per-response loss averaging assigns lower per-token weight to final answer tokens in longer reasoning traces, allocating a higher proportion of prompts to thinking mode balances supervision across both paradigms. During RL, all rollouts within a prompt group share the same operating mode.

\subsubsection{On-Policy Distillation}\label{sec:opd}
OPD provides dense, token-level teacher supervision on trajectories sampled from the student policy~\citep{agarwal2024policy}. We use a post-trained WeMM-Ecom-9B teacher for the 0.8B and 2B models, WeMM-Ecom-35B-A3B for the 4B model, and Qwen3.5-122B-A10B for the 9B and 35B-A3B models. Given a student-generated prefix $h_{i,t}=(x_i,y_{i,<t})$ of response $i$, the teacher evaluates the token $y_{i,t}$ sampled by the rollout policy $\pi_{\theta_{\mathrm{old}}}$, yielding a token-level distillation signal applied identically in both modes:
\begin{equation}
 a_{i,t}^{\mathrm{KD}}=
 \log\pi_T(y_{i,t}\mid h_{i,t})-
 \log\pi_{\theta_{\mathrm{old}}}(y_{i,t}\mid h_{i,t}),
 \label{eq:opd_signal}
\end{equation}
where $\pi_T$ denotes the teacher policy. At any fixed prefix, the expected negative signal corresponds to the reverse KL divergence from $\pi_{\theta_{\mathrm{old}}}$ to $\pi_T$. Holding $a_{i,t}^{\mathrm{KD}}$ fixed during gradient updates and letting $\rho_{i,t}=\pi_\theta(y_{i,t}\mid h_{i,t})/\pi_{\theta_{\mathrm{old}}}(y_{i,t}\mid h_{i,t})$ denote the importance ratio, we minimize the clipped actor objective over batch $\mathcal B$ and scored token positions $\mathcal R_i$:
\begin{equation}
 \mathcal L_{\mathrm{OPD}}(\theta)=
 -\frac{1}{|\mathcal B|}\sum_{i\in\mathcal B}\frac{1}{|\mathcal R_i|}
 \sum_{t\in\mathcal R_i}
 \min\!\bigl(\rho_{i,t}a_{i,t}^{\mathrm{KD}},
 \operatorname{clip}(\rho_{i,t},0.8,1.2)a_{i,t}^{\mathrm{KD}}\bigr).
 \label{eq:opd}
\end{equation}
To combine domain adaptation with the broad instruction-following priors of the backbone, we merge once after OPD and before RL by interpolating the vision and language parameters of the OPD checkpoint with the corresponding post-trained Qwen3.5 checkpoint using equal weights~\citep{dash2025aya,ilharco2022editing}.

\subsubsection{Reinforcement Learning}
\label{sec:grpo}
We further refine the models via GRPO~\citep{shao2024deepseekmath} using task-specific verifiable rewards. For a prompt $x$ from task family $\tau$, we sample $G$ responses and compute scalar rewards $r_i=R_\tau(x,y_i)$, normalizing advantages across the group:
\begin{equation}
 \widehat A_i=\frac{r_i-\bar r}
 {\operatorname{std}(r_1,\ldots,r_G)+\epsilon_A},
 \qquad \bar r=\frac1G\sum_i r_i.
 \label{eq:adv}
\end{equation}
Task rewards check predicted equivalence labels for product matching, score visual grounding and detection by weighted box IoU, and measure constraint satisfaction for instruction following (Appendix~\ref{app:commerce_rewards}). With $\ell_{i,t}=\log\pi_\theta(y_{i,t}\mid h_{i,t})$ and a frozen reference policy initialized from the pre-RL checkpoint, each online rollout batch $\mathcal B$ minimizes
\begin{equation}
 \mathcal L_{\mathrm{RL}}(\theta) =
 \frac{1}{|\mathcal B|}\sum_{i\in\mathcal B}\frac{1}{|\mathcal R_i|}
 \sum_{t\in\mathcal R_i}\left[-\widehat A_i
 \exp\!\big(\ell_{i,t}-\operatorname{sg}(\ell_{i,t})\big)
 +0.01K_{i,t}\right],
 \label{eq:rl}
\end{equation}
where $\operatorname{sg}(\cdot)$ denotes the stop-gradient operator and $K_{i,t}$ penalizes KL divergence from the reference policy (Appendix~\ref{app:rl_details}).

\subsection{Training Configuration}
CPT employs hybrid AdamW--Muon optimization~\citep{loshchilov2017decoupled,jordan2024muon}, combining AdamW updates with Muon's orthogonalized momentum. Table~\ref{tab:training_recipe} summarizes the training hyperparameters across all four stages for WeMM-Ecom-35B-A3B. At the 9B scale, CPT uses a peak learning rate of $2\times10^{-6}$ and a global batch size of 2,048 sequences. Across all five model sizes, OPD and RL use a constant learning rate of $10^{-6}$ and a global batch size of 2,048 sampled responses.

\begin{table}[!hb]
\centering
\footnotesize
\setlength{\tabcolsep}{4pt}
\renewcommand{\arraystretch}{1.04}
\caption{Training setup and hyperparameters across stages.}
\label{tab:training_recipe}
\begin{tabular*}{\linewidth}{@{\extracolsep{\fill}}>{\raggedright\arraybackslash}p{0.26\linewidth}*{4}{>{\centering\arraybackslash}p{0.148\linewidth}}@{}}
\toprule
Setting & CPT Stage 1 & CPT Stage 2 & OPD & RL/GRPO \\
\midrule
Training budget (tokens) & 370.3B & 76.7B & 0.6B & 0.2B \\
Optimization signal & Likelihood & Likelihood & Teacher scores & Task rewards \\
Optimizer & AdamW + Muon & AdamW + Muon & AdamW & AdamW \\
LR warmup steps  & 1,000 & 1,500 & 0 & 0 \\
Peak learning rate & $1.5\times10^{-6}$ & $10^{-6}$ & $10^{-6}$ & $10^{-6}$ \\
Minimum learning rate & $1.5\times10^{-7}$ & $10^{-7}$ & $10^{-6}$ & $10^{-6}$ \\
Tokens per step & 5.7M & 5.7M & -- & -- \\
Global batch size & -- & -- & 2,048 & 2,048 \\
Sequence length & 16,384 & 32,768 & 32,768 & 65,536 \\
\bottomrule
\end{tabular*}
\end{table}

\section{E-Commerce Evaluation Framework}
\label{sec:eval}

\begin{figure}[!htbp]
\centering
\includegraphics[width=0.94\linewidth]{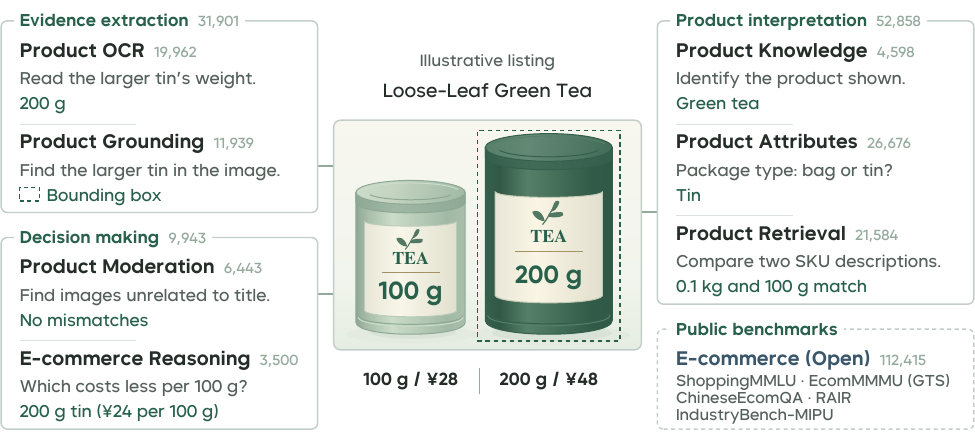}
\caption{\textbf{Overview of the eight e-commerce evaluation scenarios.} Seven in-house scenarios are organized around an illustrative product listing into evidence extraction, product interpretation, and decision making, with representative prompts and target outputs (green), complemented by five public benchmarks in E-commerce (Open). Numbers in gray denote instance counts.}
\label{fig:commerce-scenarios}
\end{figure}

Evaluating multimodal models for real-world commerce requires testing whether they can interpret complex product listings, align variants across views, enforce platform governance rules, and localize advertised items in livestream frames. We construct a comprehensive evaluation suite of 28 benchmarks---comprising 23 in-house benchmarks and five public benchmarks, totaling 207,117 evaluation instances---organized into the eight core scenarios illustrated in Figure~\ref{fig:commerce-scenarios}. Because the in-house benchmarks derive from proprietary business data, we detail their task definitions, input modalities, and scoring metrics in Appendix~\ref{app:eval_details}.

To prevent data contamination between training and business evaluation splits, we enforce strict deduplication across splits using product identifiers, cryptographic hashes for exact image duplicates, perceptual hashes for near duplicates, and visual embedding similarity.

\paragraph{Evidence extraction (Product OCR and Product Grounding).}
Product OCR (19,962 instances) evaluates fine-grained text recognition and specification reading across single product photos, multi-angle gallery views, and long promotional strips. Product Grounding (six benchmarks, 11,939 instances) assesses spatial localization of target products (Product Detection), brand logos (Logo Detection), certification marks (Qualification Detection), specification tables (Chart Detection), dimension annotations (Dimension Detection), and advertised items in livestream frames (Livestream Detection, 1,992 instances).

\paragraph{Product interpretation (Product Knowledge, Product Attributes, and Product Retrieval).}
Product Knowledge (two benchmarks, 4,598 instances) tests fine-grained entity recognition and domain knowledge across products, brands, and industry standards. Product Attributes (four benchmarks, 26,676 instances) evaluates attribute value extraction, brand attribution, and multi-choice property selection from interleaved images and structured fields. Product Retrieval (three benchmarks, 21,584 instances) benchmarks cross-item matching at three granularity levels: judging visual similarity between cropped regions (Region-level), determining whether two multimodal listings represent the same product (Product-level), and verifying exact variant equivalence from SKU specifications (SKU-level).

\paragraph{Decision making (Product Moderation and E-commerce Reasoning).}
Product Moderation (three benchmarks, 6,443 instances) evaluates rule-grounded compliance decisions, including detecting prohibited goods, enforcing category sales restrictions, and identifying image--title mismatches. E-commerce Reasoning (four benchmarks, 3,500 instances) comprises single-image analytical tasks covering chart reasoning, taxonomy classification, constrained product counting, and spatial relation reasoning, where reference answers are programmatically derived from the underlying product tables and query constraints.

\paragraph{Public benchmark coverage.}
The E-commerce (Open) scenario incorporates 112,415 instances from five public benchmarks: ShoppingMMLU and ChineseEcomQA for e-commerce concepts and operations~\citep{jin2024shopping,chen2025chineseecomqa}, RAIR for rule-aware search relevance~\citep{lu2025rair}, and EcomMMMU and IndustryBench-MIPU for multimodal product understanding and multi-image industrial attribute extraction~\citep{ling2025ecommmmu,qi2026industrybench}.
\section{Results}
\label{sec:results}

\subsection{Experimental Setup}
\label{sec:comparison}
We evaluate all five \model\ models in both thinking and non-thinking modes, using a single shared checkpoint per model scale across both operating modes and across both evaluation suites. Our primary comparison pairs each \model\ model with the corresponding post-trained Qwen3.5 baseline of the same parameter scale and mode across all 28 e-commerce benchmarks and 64 general-purpose benchmarks. 

\paragraph{Evaluation protocol.}
\label{sec:comparison_protocol}
We report the unweighted mean of primary metrics across the 28 e-commerce benchmarks and, separately, across the 64 general benchmarks on a 0--100 scale (where RefCOCO, RefCOCO+, and RefCOCOg are averaged into a single RefCOCO score and the English and Chinese splits of MMBench count as separate benchmarks). Tasks requiring open-ended semantic adjudication use Qwen3-30B-A3B-Instruct-2507~\citep{yang2025qwen3} as an independent judge, distinct from the Qwen3.5-122B-A10B verifier used during RL training. In flagship comparisons, each external baseline is reported under its best-performing evaluated operating mode, i.e., whichever of thinking and non-thinking mode scores higher for that model. Detailed scoring protocols and decoding settings are provided in Appendix~\ref{app:eval_details}.

\subsection{E-Commerce Results}
\label{sec:ecom_results}
\label{sec:final_results}

\paragraph{Flagship comparison.}
\label{sec:flagship_results}
Table~\ref{tab:results_ecom_flagship} compares WeMM-Ecom-35B-A3B against leading open-source flagships, including Qwen3.8-Flash-Next~\citep{qiu2026design}, HY4~\citep{tencent2026hy4}, MiMo-V2.6-Flash-RL~\citep{qiao2026mimo}, DeepSeek-V4.1-Flash-1017~\citep{deepseek2026v41flash1017}, and GLM-5.3-Flash~\citep{zai2026glm53flash}. Despite having $5\times$--$22\times$ fewer total parameters, WeMM-Ecom-35B-A3B achieves the highest overall average of 74.4 in non-thinking mode, outperforming the strongest baseline, Qwen3.8-Flash-Next (73.2), by about 1.2 points. 
Substantial gains in Product Retrieval and Product Grounding drive this overall lead (Table~\ref{tab:results_ecom_flagship}), which remains consistent under alternative scenario weighting schemes.

\newcommand{\resultgroupseparator}{%
  \noalign{\penalty0\kern1.05pt\hrule height .4pt\kern.9pt\hrule height .4pt\kern1.35pt\nobreak}%
}
\newcommand{\resultgroupheading}[2]{%
  \resultgroupseparator
  \multicolumn{#1}{c}{\itshape #2}\\*%
}

\begingroup
\makeatletter
\makeatother
\renewcommand{\floatpagefraction}{.9}

\paragraph{Dense models.}
\label{sec:family_comparison}
Table~\ref{tab:results_ecom_dense} reports thinking and non-thinking results for all four dense \model\ models alongside Qwen3.5-9B and the e-commerce specialist Valley3-32B~\citep{chen2026valley3}. WeMM-Ecom-9B achieves overall means of 72.5/72.1 (thinking/non-thinking), surpassing Qwen3.5-9B (67.3/65.7) by 5.2/6.4 points, Valley3-32B (64.6/68.6) by 7.9/3.5 points, and the $3\times$ larger Qwen3.8-27B~\citep{qwen2026qwen3827b} (71.8; Figure~\ref{fig:headline_results}) by 0.7 points. At the compact scales, WeMM-Ecom-4B, 2B, and 0.8B reach 68.7/68.7, 64.2/63.3, and 57.6/56.7, respectively. Across all five model sizes and both operating modes, \model\ consistently outperforms size-matched Qwen3.5, with the largest non-thinking improvements reaching +15.5 and +12.3 points at 0.8B and 2B.

\begin{table}[!htbp]
\centering
\caption{Comparison of WeMM-Ecom-35B-A3B with flagship open-source baselines on e-commerce benchmarks. Bold and underlining mark the highest and second-highest scores in each row.}\label{tab:results_ecom_flagship}
\definecolor{resultgain}{HTML}{B51E32}
\definecolor{resultloss}{HTML}{23715A}
\definecolor{resulthighlight}{HTML}{EAF5EE}
\setlength{\tabcolsep}{2pt}
\renewcommand{\arraystretch}{1.04}
\newcommand{\centerresultheader}[1]{\raisebox{\dimexpr(\depth-\height)/2\relax}{#1}}
\fontsize{8.5}{10.2}\selectfont
\begin{tabular}{@{}>{\raggedright\arraybackslash}p{0.22\linewidth}>{\columncolor{resulthighlight}\centering\arraybackslash}p{\dimexpr0.14\linewidth-2\tabcolsep\relax}>{\centering\arraybackslash}p{\dimexpr0.123\linewidth-2\tabcolsep\relax}>{\centering\arraybackslash}p{\dimexpr0.12\linewidth-2\tabcolsep\relax}>{\centering\arraybackslash}p{\dimexpr0.16\linewidth-2\tabcolsep\relax}>{\centering\arraybackslash}p{\dimexpr0.12\linewidth-2\tabcolsep\relax}>{\centering\arraybackslash}p{\dimexpr0.117\linewidth-2\tabcolsep\relax}@{}}
\toprule
{\bfseries Benchmark} & \centerresultheader{{\bfseries WeMM-Ecom}} & \centerresultheader{\shortstack{{\bfseries Qwen3.8}\\{\bfseries Flash-Next}}} & \centerresultheader{{\bfseries HY4}} & \centerresultheader{\shortstack{{\bfseries DeepSeek}\\{\bfseries V4.1-Flash-1017}}} & \centerresultheader{\shortstack{{\bfseries GLM-5.3}\\{\bfseries Flash}}} & \centerresultheader{\shortstack{{\bfseries MiMo-V2.6}\\{\bfseries Flash-RL}}} \\
\midrule[.35pt]
{\bfseries Size} & {\bfseries 35B-A3B} & {\bfseries 180B-A6B} & {\bfseries 770B-A49B} & {\bfseries 763B-A16B} & {\bfseries 320B-A18B} & {\bfseries 310B-A15B} \\
\resultgroupheading{7}{Product Knowledge}
E-commerce Knowledge & 67.1 & 78.9 & {\bfseries 85.2} & 82.4 & \underline{82.9} & 79.2 \\
Entity Recognition & 80.4 & 84.2 & {\bfseries 86.7} & 81.6 & \underline{84.9} & 76.2 \\
\resultgroupheading{7}{Product Attributes}
Attribute Selection & {\bfseries 77.1} & 75.7 & \underline{76.5} & 74.0 & 71.9 & 68.3 \\
Attribute Extraction & {\bfseries 81.8} & 78.8 & 77.9 & \underline{79.0} & 76.4 & 76.9 \\
Brand Attribution & 83.1 & {\bfseries 86.3} & \underline{84.3} & 82.5 & 80.5 & 80.8 \\
Brand Extraction & {\bfseries 83.6} & 81.2 & 80.5 & 82.7 & \underline{83.1} & 81.5 \\
\resultgroupheading{7}{Product Moderation}
Illegal Goods & 93.0 & {\bfseries 93.7} & 89.0 & {\bfseries 93.7} & \underline{93.1} & 93.0 \\
Sales Restrictions & {\bfseries 91.1} & 89.7 & \underline{90.6} & 86.6 & 88.4 & 90.2 \\
Image-Title Mismatch & 87.4 & {\bfseries 91.7} & \underline{89.7} & 87.7 & 89.5 & 84.0 \\
\resultgroupseparator
Product OCR & {\bfseries 97.8} & \underline{97.7} & 97.1 & 95.9 & 97.0 & 93.8 \\
\resultgroupheading{7}{Product Retrieval}
Region-level Retrieval & {\bfseries 85.1} & 78.6 & 76.0 & \underline{83.9} & 76.7 & 76.1 \\
Product-level Retrieval & {\bfseries 87.5} & 66.5 & 63.9 & \underline{78.3} & 61.9 & 66.1 \\
SKU-level Retrieval & {\bfseries 62.2} & 43.6 & 43.0 & \underline{49.3} & \underline{49.3} & 38.7 \\
\resultgroupheading{7}{Product Grounding}
Chart Detection & \underline{77.0} & {\bfseries 79.4} & 72.7 & 66.5 & 65.4 & 27.6 \\
Product Detection & {\bfseries 79.9} & 71.7 & \underline{73.3} & 67.1 & 67.6 & 44.3 \\
Logo Detection & \underline{58.9} & {\bfseries 59.3} & 49.2 & 42.2 & 36.0 & 6.1 \\
Qualification Detection & {\bfseries 61.8} & \underline{54.4} & 37.2 & 23.6 & 18.1 & 2.0 \\
Dimension Detection & {\bfseries 74.3} & {\bfseries 74.3} & \underline{68.1} & 55.7 & 55.4 & 26.8 \\
Livestream Detection & {\bfseries 64.7} & 47.7 & \underline{55.6} & 53.0 & 50.1 & 27.1 \\
\resultgroupheading{7}{E-commerce Reasoning}
Chart Reasoning & \underline{99.4} & {\bfseries 99.5} & 99.1 & 93.5 & 98.5 & 96.8 \\
Category Reasoning & 81.2 & \underline{85.6} & {\bfseries 85.8} & 85.2 & \underline{85.6} & 80.2 \\
Product Counting & \underline{82.2} & 79.6 & 81.2 & 69.0 & {\bfseries 83.8} & 72.8 \\
Spatial Reasoning & 31.2 & {\bfseries 35.6} & 32.6 & 19.4 & \underline{35.4} & 17.4 \\
\resultgroupheading{7}{E-commerce (Open)}
ShoppingMMLU~\citep{jin2024shopping} & 74.9 & \underline{78.0} & 76.9 & 76.0 & {\bfseries 78.3} & 76.3 \\
RAIR~\citep{lu2025rair} & 43.7 & {\bfseries 48.0} & 45.0 & 44.9 & \underline{45.9} & \underline{45.9} \\
ChineseEcomQA~\citep{chen2025chineseecomqa} & 64.2 & \underline{71.6} & {\bfseries 74.7} & 68.2 & 70.5 & 69.3 \\
EcomMMMU (GTS)~\citep{ling2025ecommmmu} & 48.0 & 50.3 & {\bfseries 54.9} & 49.3 & \underline{53.3} & 48.9 \\
IndustryBench-MIPU~\citep{qi2026industrybench} & 63.5 & 66.8 & {\bfseries 72.8} & 63.2 & \underline{69.7} & 65.3 \\
\midrule
{\bfseries Overall mean (28)} & {\bfseries 74.4} & \underline{73.2} & 72.1 & 69.1 & 69.6 & 61.1 \\
\bottomrule
\end{tabular}
\begin{papertablenotes}
  \item Each comparison model is reported in its best-performing evaluated operating mode (thinking or non-thinking).
\end{papertablenotes}
\end{table}

\begin{table}[!htbp]
\centering
\caption{Comparison of dense WeMM-Ecom models with baselines on e-commerce benchmarks. Results are thinking / non-thinking. Bold and underlining mark the highest and second-highest scores in each row. Region-level and Product-level Retrieval are scored differently in thinking and non-thinking modes; see Appendix~\ref{app:eval_details}.}\label{tab:results_ecom_dense}
\definecolor{resultgain}{HTML}{B51E32}
\definecolor{resultloss}{HTML}{23715A}
\definecolor{resulthighlight}{HTML}{EAF5EE}
\setlength{\tabcolsep}{3pt}
\renewcommand{\arraystretch}{1.04}
\newcommand{\centerresultheader}[1]{\raisebox{\dimexpr(\depth-\height)/2\relax}{#1}}
\fontsize{8.5}{10.2}\selectfont
\newcommand{\moderesult}[2]{\mbox{\fontsize{8.5}{10}\selectfont #1\,/\,#2}}
\begin{tabular}{@{}>{\raggedright\arraybackslash}p{0.22\linewidth}>{\centering\arraybackslash}p{\dimexpr0.13\linewidth-2\tabcolsep\relax}>{\centering\arraybackslash}p{\dimexpr0.13\linewidth-2\tabcolsep\relax}>{\centering\arraybackslash}p{\dimexpr0.13\linewidth-2\tabcolsep\relax}>{\columncolor{resulthighlight}\centering\arraybackslash}p{\dimexpr0.13\linewidth-2\tabcolsep\relax}>{\centering\arraybackslash}p{\dimexpr0.13\linewidth-2\tabcolsep\relax}>{\centering\arraybackslash}p{\dimexpr0.13\linewidth-2\tabcolsep\relax}@{}}
\toprule
{\bfseries Benchmark} & \multicolumn{4}{c}{\centerresultheader{{\bfseries WeMM-Ecom}}} & \centerresultheader{{\bfseries Qwen3.5}} & \centerresultheader{{\bfseries Valley3}} \\
\midrule[.35pt]
{\bfseries Size} & {\bfseries 0.8B} & {\bfseries 2B} & {\bfseries 4B} & {\bfseries 9B} & {\bfseries 9B} & {\bfseries 32B} \\
\resultgroupheading{7}{Product Knowledge}
E-commerce Knowledge & \moderesult{26.7}{30.8} & \moderesult{36.4}{38.6} & \moderesult{50.9}{51.4} & \moderesult{58.7}{59.4} & \moderesult{57.5}{55.8} & \moderesult{{\bfseries 63.4}}{\underline{62.8}} \\
Entity Recognition & \moderesult{53.1}{66.9} & \moderesult{74.4}{76.6} & \moderesult{78.4}{78.4} & \moderesult{\underline{80.2}}{77.3} & \moderesult{79.8}{75.6} & \moderesult{79.3}{{\bfseries 80.6}} \\
\resultgroupheading{7}{Product Attributes}
Attribute Selection & \moderesult{66.2}{62.6} & \moderesult{71.1}{71.3} & \moderesult{73.3}{\underline{74.1}} & \moderesult{73.6}{{\bfseries 74.5}} & \moderesult{71.0}{70.1} & \moderesult{69.2}{71.9} \\
Attribute Extraction & \moderesult{71.8}{75.2} & \moderesult{77.6}{79.2} & \moderesult{79.9}{80.4} & \moderesult{\underline{80.8}}{{\bfseries 81.1}} & \moderesult{77.0}{73.6} & \moderesult{74.1}{78.6} \\
Brand Attribution & \moderesult{79.1}{79.2} & \moderesult{82.6}{\underline{83.3}} & \moderesult{81.8}{80.8} & \moderesult{{\bfseries 83.5}}{81.0} & \moderesult{79.8}{73.6} & \moderesult{77.0}{80.3} \\
Brand Extraction & \moderesult{80.1}{80.8} & \moderesult{81.0}{82.5} & \moderesult{80.8}{83.2} & \moderesult{81.2}{{\bfseries 85.1}} & \moderesult{81.8}{81.8} & \moderesult{83.1}{\underline{83.8}} \\
\resultgroupheading{7}{Product Moderation}
Illegal Goods & \moderesult{83.8}{84.2} & \moderesult{88.4}{87.4} & \moderesult{91.5}{91.5} & \moderesult{{\bfseries 91.9}}{91.7} & \moderesult{91.0}{88.4} & \moderesult{\underline{91.8}}{90.3} \\
Sales Restrictions & \moderesult{82.6}{77.5} & \moderesult{84.3}{83.1} & \moderesult{89.3}{90.0} & \moderesult{89.9}{{\bfseries 90.8}} & \moderesult{89.7}{83.3} & \moderesult{\underline{90.4}}{88.2} \\
Image-Title Mismatch & \moderesult{77.2}{38.5} & \moderesult{82.7}{62.3} & \moderesult{\underline{88.3}}{82.2} & \moderesult{85.7}{85.6} & \moderesult{80.2}{70.1} & \moderesult{{\bfseries 89.1}}{86.7} \\
\resultgroupseparator
Product OCR & \moderesult{93.3}{93.8} & \moderesult{96.3}{96.2} & \moderesult{97.1}{\underline{97.5}} & \moderesult{{\bfseries 97.6}}{97.3} & \moderesult{97.2}{96.4} & \moderesult{95.1}{96.1} \\
\resultgroupheading{7}{Product Retrieval}
Region-level Retrieval & \moderesult{74.2}{83.5} & \moderesult{78.8}{\underline{84.4}} & \moderesult{80.2}{84.3} & \moderesult{80.7}{{\bfseries 85.6}} & \moderesult{76.9}{80.7} & \moderesult{73.4}{81.6} \\
Product-level Retrieval & \moderesult{64.3}{85.3} & \moderesult{60.9}{{\bfseries 88.5}} & \moderesult{64.7}{85.3} & \moderesult{61.1}{\underline{85.8}} & \moderesult{61.6}{79.9} & \moderesult{67.9}{83.6} \\
SKU-level Retrieval & \moderesult{53.7}{63.1} & \moderesult{62.4}{{\bfseries 64.8}} & \moderesult{58.3}{\underline{64.0}} & \moderesult{61.9}{61.4} & \moderesult{52.1}{32.0} & \moderesult{33.0}{43.2} \\
\resultgroupheading{7}{Product Grounding}
Chart Detection & \moderesult{52.4}{50.6} & \moderesult{64.4}{55.7} & \moderesult{61.4}{66.9} & \moderesult{{\bfseries 73.4}}{\underline{73.3}} & \moderesult{60.8}{63.6} & \moderesult{69.4}{65.4} \\
Product Detection & \moderesult{74.8}{75.1} & \moderesult{68.6}{72.3} & \moderesult{76.4}{78.5} & \moderesult{\underline{79.2}}{{\bfseries 80.4}} & \moderesult{70.0}{72.3} & \moderesult{66.0}{71.8} \\
Logo Detection & \moderesult{45.5}{41.8} & \moderesult{44.2}{43.1} & \moderesult{\underline{54.1}}{51.1} & \moderesult{{\bfseries 59.1}}{54.0} & \moderesult{44.5}{43.4} & \moderesult{41.5}{42.0} \\
Qualification Detection & \moderesult{16.6}{19.7} & \moderesult{24.7}{25.1} & \moderesult{27.9}{35.5} & \moderesult{\underline{49.0}}{{\bfseries 51.4}} & \moderesult{22.6}{37.2} & \moderesult{34.3}{40.2} \\
Dimension Detection & \moderesult{30.5}{42.2} & \moderesult{44.4}{38.7} & \moderesult{40.5}{36.8} & \moderesult{\underline{77.2}}{{\bfseries 85.9}} & \moderesult{55.9}{75.2} & \moderesult{41.1}{73.2} \\
Livestream Detection & \moderesult{55.9}{58.5} & \moderesult{59.9}{\underline{60.9}} & \moderesult{59.2}{{\bfseries 61.8}} & \moderesult{58.6}{60.2} & \moderesult{48.5}{52.0} & \moderesult{40.9}{47.1} \\
\resultgroupheading{7}{E-commerce Reasoning}
Chart Reasoning & \moderesult{77.6}{65.9} & \moderesult{96.3}{94.4} & \moderesult{\underline{99.2}}{98.3} & \moderesult{{\bfseries 99.4}}{95.9} & \moderesult{98.7}{98.4} & \moderesult{73.6}{89.5} \\
Category Reasoning & \moderesult{77.6}{76.6} & \moderesult{80.2}{80.4} & \moderesult{\underline{83.4}}{82.2} & \moderesult{{\bfseries 84.2}}{82.4} & \moderesult{83.0}{80.6} & \moderesult{81.6}{81.6} \\
Product Counting & \moderesult{38.0}{34.4} & \moderesult{63.2}{56.0} & \moderesult{\underline{79.8}}{70.8} & \moderesult{{\bfseries 80.4}}{68.6} & \moderesult{77.4}{70.2} & \moderesult{60.6}{74.2} \\
Spatial Reasoning & \moderesult{3.8}{1.2} & \moderesult{13.4}{10.6} & \moderesult{\underline{32.4}}{24.4} & \moderesult{{\bfseries 38.2}}{21.6} & \moderesult{29.2}{20.6} & \moderesult{16.4}{17.4} \\
\resultgroupheading{7}{E-commerce (Open)}
ShoppingMMLU~\citep{jin2024shopping} & \moderesult{61.1}{50.2} & \moderesult{70.5}{60.6} & \moderesult{74.9}{69.7} & \moderesult{{\bfseries 76.6}}{72.5} & \moderesult{\underline{76.0}}{69.4} & \moderesult{73.3}{73.8} \\
RAIR~\citep{lu2025rair} & \moderesult{37.0}{23.8} & \moderesult{36.1}{36.6} & \moderesult{40.5}{36.1} & \moderesult{44.3}{42.7} & \moderesult{\underline{44.6}}{41.5} & \moderesult{{\bfseries 48.7}}{44.5} \\
ChineseEcomQA~\citep{chen2025chineseecomqa} & \moderesult{50.6}{45.4} & \moderesult{57.6}{46.9} & \moderesult{61.1}{61.1} & \moderesult{\underline{64.3}}{60.0} & \moderesult{61.4}{54.8} & \moderesult{59.8}{{\bfseries 66.9}} \\
EcomMMMU (GTS)~\citep{ling2025ecommmmu} & \moderesult{40.5}{41.5} & \moderesult{41.1}{41.7} & \moderesult{50.6}{50.4} & \moderesult{\underline{51.5}}{{\bfseries 52.0}} & \moderesult{48.6}{42.0} & \moderesult{46.8}{48.5} \\
IndustryBench-MIPU~\citep{qi2026industrybench} & \moderesult{44.9}{39.7} & \moderesult{57.6}{51.9} & \moderesult{{\bfseries 67.1}}{57.0} & \moderesult{66.5}{61.4} & \moderesult{\underline{66.6}}{56.1} & \moderesult{{\bfseries 67.1}}{57.4} \\
\midrule
{\bfseries Overall mean (28)} & \moderesult{57.6}{56.7} & \moderesult{64.2}{63.3} & \moderesult{68.7}{68.7} & \moderesult{{\bfseries 72.5}}{\underline{72.1}} & \moderesult{67.3}{65.7} & \moderesult{64.6}{68.6} \\
\bottomrule
\end{tabular}
\end{table}

\subsection{Analysis of E-Commerce Gains}
\label{sec:mode_results}
\label{sec:qualitative_protocol}
Fine-grained visual localization and cross-listing variant matching account for the largest capability leaps on the e-commerce suite. At the 9B scale, \model\ improves upon Qwen3.5-9B across all eight scenario means in thinking mode and seven of eight in non-thinking mode. Product Grounding improves by 15.7 points in thinking mode and 10.3 points in non-thinking mode, while Product Retrieval gains 13.4 points in non-thinking mode---directly reflecting the bounding-box localization and multi-level equivalence supervision introduced in Section~\ref{sec:ecom_data}.

\paragraph{Broad improvements beyond grounding.}
To confirm that \model's advantage extends well beyond spatial localization, we re-evaluate the suite under alternative subsets and weighting schemes.
Excluding all six Product Grounding benchmarks retains clear positive gains over size-matched Qwen3.5 across all ten size--mode pairs---including non-thinking improvements of +11.1 and +8.7 points at 0.8B and 2B---and leaves WeMM-Ecom-35B-A3B with the highest mean among the models in Table~\ref{tab:results_ecom_flagship} (75.7 for WeMM-Ecom-35B-A3B versus 75.5 for Qwen3.8-Flash-Next, 75.6 for HY4, and 73.9 for DeepSeek-V4.1-Flash-1017, keeping each model in its Table~\ref{tab:results_ecom_flagship} operating mode). 
Likewise, \model\ maintains positive gains across all ten size--mode comparisons when weighting all eight scenarios equally (+2.1 to +14.8 points, which also preserves the flagship ranking), when omitting any single scenario (at least +0.9 points), or when excluding the two probability-readout retrieval sets (+3.1 to +17.8 points).

\paragraph{Scenario-level trade-offs and format robustness.}
While overall e-commerce performance improves across scales, gains vary by task family: Qwen3.5-35B-A3B is slightly higher on Product Moderation in thinking mode, and Qwen3.5-9B slightly leads WeMM-Ecom-9B on E-commerce Reasoning in non-thinking mode. For structured localization tasks such as Chart Detection, our evaluation parser accepts class predictions under both \texttt{category} and \texttt{label} keys so that comparisons measure localization quality rather than JSON key conventions (Appendix~\ref{app:grounding_protocol}). Even under this schema-compatible parser, WeMM-Ecom-35B-A3B outperforms Qwen3.5-35B-A3B on Chart Detection by 8.1 points in thinking mode and 13.0 points in non-thinking mode.

\endgroup

\subsection{General Benchmarks}
\label{sec:general_eval}
To assess general-purpose perception and reasoning, we evaluate models across 64 public benchmarks comprising 54 multimodal benchmarks---such as MMBench, MMMU, and MathVista~\citep{liu2024mmbench,yue2024mmmu,lu2024mathvista}---and 10 text-only benchmarks, including MMLU-Pro and IFEval~\citep{wang2024mmlu,zhou2023instruction}.

\paragraph{Comparison with state-of-the-art models.}
Table~\ref{tab:results_general_flagship} compares WeMM-Ecom-35B-A3B with Qwen3.5-35B-A3B~\citep{qwenteam2026qwen35}, Qwen3.6-35B-A3B~\citep{qwen2026qwen3635ba3b}, MiniCPM-V-4.7-35B-A3B~\citep{yu2026minicpm}, and Gemma4-26B-A4B-it~\citep{team2026gemma}. In thinking mode, WeMM-Ecom-35B-A3B achieves the highest overall average of 76.4, edging out both Qwen3.5-35B-A3B (76.2) and Qwen3.6-35B-A3B (76.3). In the corresponding non-thinking comparison, WeMM-Ecom-35B-A3B scores 72.6, outperforming Qwen3.5-35B-A3B (71.0) by 1.6 points.

\begingroup
\captionsetup{skip=5pt}
\definecolor{resultgain}{HTML}{B51E32}
\definecolor{resultloss}{HTML}{23715A}
\definecolor{resulthighlight}{HTML}{EAF5EE}
\setlength{\tabcolsep}{3pt}
\setlength{\LTleft}{\fill}\setlength{\LTright}{\fill}
\setlength{\LTcapwidth}{\linewidth}
\setlength{\LTpost}{7pt}
\renewcommand{\arraystretch}{1.00}
\newcommand{\centerresultheader}[1]{\raisebox{\dimexpr(\depth-\height)/2\relax}{#1}}
\fontsize{8.5}{10.0}\selectfont
\newcommand{\moderesult}[2]{\mbox{\fontsize{8.5}{10}\selectfont #1\,/\,#2}}
\begin{longtable}{@{}>{\raggedright\arraybackslash}p{0.235\linewidth}>{\columncolor{resulthighlight}\centering\arraybackslash}p{\dimexpr0.149\linewidth-2\tabcolsep\relax}>{\centering\arraybackslash}p{\dimexpr0.149\linewidth-2\tabcolsep\relax}>{\centering\arraybackslash}p{\dimexpr0.149\linewidth-2\tabcolsep\relax}>{\centering\arraybackslash}p{\dimexpr0.169\linewidth-2\tabcolsep\relax}>{\centering\arraybackslash}p{\dimexpr0.149\linewidth-2\tabcolsep\relax}@{}}
\caption{Comparison of WeMM-Ecom-35B-A3B with comparable baselines on general benchmarks. Results are thinking / non-thinking. Bold and underlining mark the highest and second-highest scores in each row.}\label{tab:results_general_flagship}\\
\toprule
{\bfseries Benchmark} & {\bfseries WeMM-Ecom} & {\bfseries Qwen3.5} & {\bfseries Qwen3.6} & \mbox{\bfseries MiniCPM-V-4.7} & {\bfseries Gemma4} \\
\midrule[.35pt]
{\bfseries Size} & {\bfseries 35B-A3B} & {\bfseries 35B-A3B} & {\bfseries 35B-A3B} & {\bfseries 35B-A3B} & {\bfseries 26B-A4B} \\
\endfirsthead
\multicolumn{6}{l}{\textbf{Table \thetable\ (continued)}} \\
\toprule
{\bfseries Benchmark} & {\bfseries WeMM-Ecom} & {\bfseries Qwen3.5} & {\bfseries Qwen3.6} & \mbox{\bfseries MiniCPM-V-4.7} & {\bfseries Gemma4} \\
\midrule[.35pt]
{\bfseries Size} & {\bfseries 35B-A3B} & {\bfseries 35B-A3B} & {\bfseries 35B-A3B} & {\bfseries 35B-A3B} & {\bfseries 26B-A4B} \\
\endhead
\midrule
\multicolumn{6}{r}{\footnotesize Continued on next page} \\
\endfoot
\bottomrule
\endlastfoot
\resultgroupheading{6}{STEM and Multimodal Reasoning}
CMMMU (val)~\citep{zhang2024cmmmu} & \moderesult{\underline{73.7}}{69.8} & \moderesult{73.6}{69.1} & \moderesult{{\bfseries 73.8}}{67.1} & \moderesult{70.6}{70.0} & \moderesult{59.3}{60.9} \\*
CMMU (val, MCQ)~\citep{he2024cmmu} & \moderesult{\underline{93.2}}{91.9} & \moderesult{{\bfseries 93.4}}{91.4} & \moderesult{92.3}{91.1} & \moderesult{92.7}{92.8} & \moderesult{71.7}{74.8} \\
CSVQA~\citep{jian2025csvqa} & \moderesult{\underline{83.5}}{73.6} & \moderesult{83.3}{75.3} & \moderesult{{\bfseries 84.2}}{74.2} & \moderesult{77.3}{81.8} & \moderesult{43.5}{48.5} \\
DynaMath~\citep{zou2025dynamath} & \moderesult{{\bfseries 84.6}}{\underline{83.5}} & \moderesult{81.8}{79.8} & \moderesult{81.5}{77.6} & \moderesult{81.6}{81.7} & \moderesult{64.6}{72.9} \\
LogicVista~\citep{xiao2024logicvista} & \moderesult{\underline{78.0}}{74.0} & \moderesult{{\bfseries 79.0}}{72.0} & \moderesult{\underline{78.0}}{71.0} & \moderesult{69.0}{72.0} & \moderesult{63.0}{67.0} \\
MATH-Vision~\citep{wang2024measuring} & \moderesult{79.6}{72.4} & \moderesult{\underline{80.9}}{72.6} & \moderesult{{\bfseries 81.7}}{73.3} & \moderesult{66.0}{72.2} & \moderesult{64.2}{65.4} \\
MathVista (mini)~\citep{lu2024mathvista} & \moderesult{{\bfseries 87.5}}{85.7} & \moderesult{85.8}{83.7} & \moderesult{86.2}{84.3} & \moderesult{85.5}{\underline{86.5}} & \moderesult{75.9}{76.6} \\
MMMU~\citep{yue2024mmmu} & \moderesult{80.1}{79.3} & \moderesult{\underline{80.8}}{78.6} & \moderesult{{\bfseries 82.9}}{79.1} & \moderesult{78.3}{79.1} & \moderesult{74.4}{73.1} \\
MMMU-Pro~\citep{yue2025mmmu} & \moderesult{\underline{74.1}}{69.2} & \moderesult{{\bfseries 74.5}}{69.5} & \moderesult{{\bfseries 74.5}}{69.0} & \moderesult{62.7}{66.1} & \moderesult{47.3}{56.1} \\
VisualPuzzles~\citep{song2025visualpuzzles} & \moderesult{{\bfseries 61.8}}{54.8} & \moderesult{60.3}{53.5} & \moderesult{\underline{61.6}}{53.3} & \moderesult{52.7}{56.3} & \moderesult{46.0}{53.0} \\
VisuLogic~\citep{xu2026visulogic} & \moderesult{31.7}{32.6} & \moderesult{32.3}{31.2} & \moderesult{{\bfseries 33.3}}{28.1} & \moderesult{29.5}{\underline{32.7}} & \moderesult{11.2}{27.7} \\
ZeroBench (sub)~\citep{roberts2025zerobench} & \moderesult{\underline{35.0}}{31.7} & \moderesult{34.7}{30.2} & \moderesult{{\bfseries 36.8}}{29.9} & \moderesult{25.5}{26.7} & \moderesult{11.7}{21.0} \\
\resultgroupheading{6}{General VQA}
CCBench~\citep{opencompassccbench} & \moderesult{78.2}{79.8} & \moderesult{76.1}{77.5} & \moderesult{76.5}{73.9} & \moderesult{{\bfseries 82.4}}{\underline{81.2}} & \moderesult{61.4}{61.6} \\*
ChineseSimpleVQA~\citep{gu2025see} & \moderesult{{\bfseries 58.8}}{56.1} & \moderesult{\underline{56.4}}{54.2} & \moderesult{55.2}{54.7} & \moderesult{52.1}{52.0} & \moderesult{53.1}{47.4} \\
CII-Bench~\citep{zhang2025can} & \moderesult{{\bfseries 70.9}}{67.0} & \moderesult{67.6}{67.1} & \moderesult{\underline{69.8}}{68.6} & \moderesult{66.6}{68.8} & \moderesult{67.1}{69.6} \\
\mbox{MMBench\textsubscript{EN-DEV-v1.1}~\citep{liu2024mmbench}} & \moderesult{90.5}{89.3} & \moderesult{90.6}{88.5} & \moderesult{90.6}{88.4} & \moderesult{\underline{91.3}}{{\bfseries 92.0}} & \moderesult{86.1}{86.7} \\*
\mbox{MMBench\textsubscript{CN-DEV-v1.1}~\citep{liu2024mmbench}} & \moderesult{\underline{90.1}}{89.5} & \moderesult{\underline{90.1}}{88.6} & \moderesult{89.6}{87.2} & \moderesult{{\bfseries 91.3}}{88.9} & \moderesult{85.6}{84.9} \\
MMStar~\citep{chen2024we} & \moderesult{{\bfseries 79.1}}{78.1} & \moderesult{\underline{79.0}}{77.7} & \moderesult{78.9}{78.0} & \moderesult{77.1}{77.5} & \moderesult{73.0}{72.2} \\
MM-Vet~\citep{yu2023mm} & \moderesult{{\bfseries 89.4}}{86.0} & \moderesult{85.9}{82.5} & \moderesult{\underline{87.1}}{82.1} & \moderesult{83.4}{84.3} & \moderesult{79.1}{76.6} \\
PerceptionBench~\citep{lin2026perceptionbench} & \moderesult{41.5}{38.4} & \moderesult{{\bfseries 43.6}}{37.3} & \moderesult{\underline{42.9}}{36.6} & \moderesult{34.2}{35.8} & \moderesult{32.4}{34.4} \\
RealWorldQA~\citep{xai2024realworldqa} & \moderesult{83.0}{78.7} & \moderesult{{\bfseries 84.7}}{77.3} & \moderesult{\underline{83.1}}{81.7} & \moderesult{76.3}{76.6} & \moderesult{65.8}{67.2} \\
SimpleVQA~\citep{cheng2025simplevqa} & \moderesult{{\bfseries 67.6}}{60.8} & \moderesult{\underline{62.9}}{59.5} & \moderesult{60.7}{58.3} & \moderesult{61.0}{58.4} & \moderesult{56.4}{54.0} \\
WorldVQA~\citep{zhou2026worldvqa} & \moderesult{23.4}{24.0} & \moderesult{\underline{26.8}}{23.1} & \moderesult{{\bfseries 27.2}}{24.8} & \moderesult{15.7}{16.1} & \moderesult{14.9}{14.0} \\
\resultgroupheading{6}{Documents and Charts}
AI2D (test)~\citep{kembhavi2016diagram} & \moderesult{89.4}{89.3} & \moderesult{\underline{90.4}}{88.9} & \moderesult{{\bfseries 91.0}}{89.4} & \moderesult{90.3}{89.6} & \moderesult{83.9}{84.7} \\*
CC-OCR~\citep{yang2025cc} & \moderesult{85.1}{{\bfseries 86.3}} & \moderesult{85.5}{\underline{86.2}} & \moderesult{85.2}{86.1} & \moderesult{77.3}{81.6} & \moderesult{65.0}{76.0} \\
ChartQAPro~\citep{masry2025chartqapro} & \moderesult{{\bfseries 68.9}}{67.2} & \moderesult{\underline{68.2}}{66.3} & \moderesult{\underline{68.2}}{64.4} & \moderesult{64.4}{65.6} & \moderesult{58.9}{63.0} \\
ChartQA (test)~\citep{masry2022chartqa} & \moderesult{\underline{91.0}}{88.4} & \moderesult{{\bfseries 91.2}}{86.8} & \moderesult{\underline{91.0}}{87.0} & \moderesult{89.4}{86.9} & \moderesult{87.5}{83.4} \\
CharXiv (RQ)~\citep{wang2024charxiv} & \moderesult{\underline{76.8}}{70.4} & \moderesult{{\bfseries 77.8}}{69.9} & \moderesult{75.0}{69.6} & \moderesult{67.9}{68.3} & \moderesult{63.6}{62.8} \\
DocVQA (val)~\citep{mathew2021docvqa} & \moderesult{{\bfseries 97.7}}{96.9} & \moderesult{\underline{97.5}}{96.6} & \moderesult{97.3}{96.0} & \moderesult{96.5}{94.6} & \moderesult{89.9}{90.3} \\
InfographicVQA (val)~\citep{mathew2022infographicvqa} & \moderesult{92.8}{91.2} & \moderesult{\underline{93.0}}{90.7} & \moderesult{{\bfseries 93.2}}{91.3} & \moderesult{85.8}{84.9} & \moderesult{70.7}{67.0} \\
OCRBench~\citep{liu2024ocrbench} & \moderesult{{\bfseries 90.9}}{\underline{90.6}} & \moderesult{88.7}{89.5} & \moderesult{88.5}{90.1} & \moderesult{87.5}{89.1} & \moderesult{82.9}{83.6} \\
OCRBench v2~\citep{fu2026ocrbench} & \moderesult{64.6}{63.9} & \moderesult{65.5}{64.1} & \moderesult{{\bfseries 66.8}}{\underline{66.5}} & \moderesult{49.5}{61.8} & \moderesult{39.1}{44.3} \\
OmniDocBench~\citep{ouyang2025omnidocbench} & \moderesult{90.6}{{\bfseries 92.7}} & \moderesult{89.6}{88.6} & \moderesult{91.5}{90.3} & \moderesult{77.6}{\underline{92.0}} & \moderesult{3.1}{27.0} \\
WildDoc~\citep{wang2025wilddoc} & \moderesult{68.6}{70.9} & \moderesult{{\bfseries 77.0}}{74.8} & \moderesult{\underline{76.8}}{74.4} & \moderesult{67.3}{70.5} & \moderesult{43.3}{48.7} \\
\resultgroupheading{6}{Grounding and Counting}
CountBenchQA~\citep{beyer2024paligemma} & \moderesult{\underline{96.5}}{91.0} & \moderesult{{\bfseries 96.7}}{92.1} & \moderesult{96.1}{92.1} & \moderesult{92.9}{90.4} & \moderesult{95.1}{85.7} \\*
CountQA~\citep{tamarapalli2025countqa} & \moderesult{{\bfseries 70.2}}{43.9} & \moderesult{\underline{69.4}}{40.1} & \moderesult{67.3}{40.1} & \moderesult{35.5}{31.7} & \moderesult{39.3}{26.6} \\
MIHBench~\citep{li2025mihbench} & \moderesult{{\bfseries 83.1}}{79.1} & \moderesult{\underline{82.4}}{60.0} & \moderesult{81.6}{63.9} & \moderesult{81.3}{81.5} & \moderesult{75.8}{44.0} \\
ODInW13~\citep{li2022grounded} & \moderesult{48.6}{44.5} & \moderesult{45.0}{48.8} & \moderesult{{\bfseries 50.8}}{\underline{50.4}} & \moderesult{26.3}{27.8} & \moderesult{35.3}{33.0} \\
PixMo-Count~\citep{deitke2025molmo} & \moderesult{\underline{89.9}}{75.4} & \moderesult{{\bfseries 90.1}}{75.8} & \moderesult{89.7}{73.5} & \moderesult{71.0}{69.3} & \moderesult{81.9}{62.4} \\
RefCOCO (avg)~\citep{yu2016modeling,mao2016generation} & \moderesult{88.6}{88.9} & \moderesult{88.4}{\underline{89.2}} & \moderesult{88.1}{{\bfseries 89.3}} & \moderesult{88.7}{88.9} & \moderesult{79.1}{79.0} \\
Ref-L4 (val)~\citep{chen2025revisiting} & \moderesult{\underline{90.0}}{{\bfseries 90.5}} & \moderesult{89.2}{88.1} & \moderesult{88.8}{88.3} & \moderesult{77.8}{86.5} & \moderesult{76.4}{86.0} \\
TallyQA~\citep{acharya2019tallyqa} & \moderesult{{\bfseries 83.1}}{\underline{82.6}} & \moderesult{80.3}{81.5} & \moderesult{79.7}{81.3} & \moderesult{81.2}{80.6} & \moderesult{78.8}{79.0} \\
\resultgroupheading{6}{2D/3D Spatial Understanding}
CV-Bench~\citep{tong2024cambrian} & \moderesult{{\bfseries 88.9}}{\underline{88.4}} & \moderesult{88.0}{87.8} & \moderesult{87.7}{87.9} & \moderesult{86.8}{86.5} & \moderesult{84.2}{80.0} \\*
EmbSpatial-Bench~\citep{du2024embspatial} & \moderesult{83.7}{78.9} & \moderesult{\underline{83.9}}{79.3} & \moderesult{{\bfseries 84.1}}{79.3} & \moderesult{82.7}{82.2} & \moderesult{78.5}{70.3} \\*
ERQA~\citep{team2025gemini} & \moderesult{56.5}{48.8} & \moderesult{{\bfseries 61.0}}{50.3} & \moderesult{\underline{58.5}}{52.3} & \moderesult{55.0}{49.8} & \moderesult{47.0}{45.3} \\*
MMSI-Bench~\citep{yang2026mmsi} & \moderesult{39.1}{32.8} & \moderesult{\underline{40.2}}{32.2} & \moderesult{{\bfseries 41.4}}{32.4} & \moderesult{29.4}{35.1} & \moderesult{27.1}{28.8} \\*
OmniSpatial~\citep{jia2026omnispatial} & \moderesult{{\bfseries 61.0}}{55.5} & \moderesult{\underline{60.6}}{53.0} & \moderesult{59.8}{54.1} & \moderesult{53.5}{53.6} & \moderesult{49.3}{44.9} \\*
SpatialMQA~\citep{liu2025can} & \moderesult{\underline{58.5}}{47.5} & \moderesult{{\bfseries 58.9}}{45.3} & \moderesult{58.1}{47.1} & \moderesult{50.9}{47.5} & \moderesult{57.8}{44.5} \\
\resultgroupheading{6}{Multi-Image Understanding}
BLINK~\citep{fu2024blink} & \moderesult{\underline{69.7}}{68.6} & \moderesult{\underline{69.7}}{66.1} & \moderesult{{\bfseries 70.4}}{65.6} & \moderesult{66.4}{67.0} & \moderesult{58.3}{63.0} \\*
Mantis-Eval~\citep{jiang2024mantis} & \moderesult{\underline{88.5}}{86.6} & \moderesult{87.6}{84.8} & \moderesult{{\bfseries 90.3}}{87.6} & \moderesult{87.6}{86.6} & \moderesult{83.0}{74.7} \\
MMIU~\citep{meng2025mmiu} & \moderesult{66.2}{66.2} & \moderesult{\underline{67.6}}{63.8} & \moderesult{{\bfseries 70.1}}{64.1} & \moderesult{64.4}{63.5} & \moderesult{53.4}{50.1} \\
MuirBench~\citep{wang2025muirbench} & \moderesult{70.9}{71.0} & \moderesult{73.2}{62.0} & \moderesult{{\bfseries 74.7}}{\underline{73.7}} & \moderesult{72.2}{65.5} & \moderesult{57.8}{60.0} \\
\resultgroupheading{6}{Instruction Following}
CFBench~\citep{zhang2025cfbench} & \moderesult{\underline{86.2}}{84.4} & \moderesult{85.2}{84.9} & \moderesult{{\bfseries 87.0}}{84.1} & \moderesult{\underline{86.2}}{85.6} & \moderesult{78.6}{82.8} \\*
IFBench~\citep{pyatkin2026generalizing} & \moderesult{{\bfseries 75.7}}{51.0} & \moderesult{\underline{71.3}}{43.3} & \moderesult{65.3}{35.7} & \moderesult{55.3}{42.3} & \moderesult{64.7}{43.0} \\
IFEval~\citep{zhou2023instruction} & \moderesult{{\bfseries 94.3}}{86.7} & \moderesult{\underline{93.7}}{85.4} & \moderesult{91.7}{83.2} & \moderesult{91.1}{88.4} & \moderesult{87.3}{87.8} \\
MIA-Bench~\citep{qian2025mia} & \moderesult{{\bfseries 87.2}}{83.8} & \moderesult{\underline{86.8}}{82.8} & \moderesult{{\bfseries 87.2}}{85.1} & \moderesult{83.7}{86.2} & \moderesult{83.1}{81.3} \\
MM-IFEval~\citep{ding2025mm} & \moderesult{\underline{82.5}}{74.3} & \moderesult{{\bfseries 82.8}}{65.2} & \moderesult{80.9}{66.3} & \moderesult{69.9}{70.5} & \moderesult{69.8}{69.6} \\
\resultgroupheading{6}{Text-Centric Tasks}
C-Eval~\citep{huang2023c} & \moderesult{{\bfseries 91.4}}{89.7} & \moderesult{\underline{91.1}}{90.4} & \moderesult{90.8}{90.7} & \moderesult{90.1}{89.9} & \moderesult{79.4}{74.9} \\*
GPQA (Diamond)~\citep{rein2023gpqa} & \moderesult{81.8}{77.8} & \moderesult{81.3}{{\bfseries 86.4}} & \moderesult{81.8}{\underline{84.9}} & \moderesult{74.8}{76.8} & \moderesult{64.7}{63.6} \\
HMMT (Nov.\ 2025)~\citep{dekoninck2026beyond} & \moderesult{\underline{87.5}}{84.2} & \moderesult{86.7}{69.2} & \moderesult{{\bfseries 88.3}}{67.5} & \moderesult{78.3}{65.8} & \moderesult{66.7}{64.2} \\
IMO-AnswerBench (v2)~\citep{luong2025towards} & \moderesult{73.3}{62.5} & \moderesult{\underline{74.0}}{58.3} & \moderesult{{\bfseries 75.3}}{59.8} & \moderesult{67.3}{56.5} & \moderesult{52.5}{50.8} \\
MMLU-Pro~\citep{wang2024mmlu} & \moderesult{\underline{85.3}}{84.1} & \moderesult{{\bfseries 85.7}}{84.3} & \moderesult{\underline{85.3}}{84.5} & \moderesult{84.5}{84.6} & \moderesult{65.7}{80.9} \\
MMLU-Redux~\citep{gema2025we} & \moderesult{{\bfseries 93.5}}{91.3} & \moderesult{{\bfseries 93.5}}{88.8} & \moderesult{\underline{93.4}}{{\bfseries 93.5}} & \moderesult{93.3}{\underline{93.4}} & \moderesult{86.0}{90.2} \\
SuperGPQA~\citep{du2026supergpqa} & \moderesult{64.3}{61.1} & \moderesult{\underline{64.8}}{62.5} & \moderesult{{\bfseries 65.9}}{63.7} & \moderesult{60.5}{61.3} & \moderesult{44.0}{52.1} \\
\midrule
{\bfseries Overall mean (64)} & \moderesult{{\bfseries 76.4}}{72.6} & \moderesult{76.2}{71.0} & \moderesult{\underline{76.3}}{71.2} & \moderesult{70.5}{70.8} & \moderesult{62.4}{62.4} \\
\end{longtable}
\endgroup

\paragraph{Dense models.}
Table~\ref{tab:results_general_dense} presents thinking and non-thinking results for our four dense models alongside Qwen3.5-9B and GLM-4.6V-Flash~\citep{hong2025glm}. WeMM-Ecom-9B averages 74.2/69.7 (thinking/non-thinking), outperforming Qwen3.5-9B (73.4/67.6) by 0.8/2.1 points and GLM-4.6V-Flash (63.3/57.9) by 10.8/11.8 points. At the compact scales (4B, 2B, and 0.8B), \model\ surpasses size-matched Qwen3.5 in thinking mode by 0.4, 4.5, and 6.9 points, respectively. As shown in Figure~\ref{fig:headline_results}, WeMM-Ecom-2B ranks first among lightweight models in non-thinking mode.

\paragraph{Robustness against published baselines.}
\label{sec:general_baseline_checks}
As a stringent external check, we also compare \model\ against an upper-bound Qwen3.5 baseline constructed by taking the higher of our local evaluation and the official Qwen3.5 model-card score~\citep{qwenteam2026qwen35} on every comparable benchmark. Even against these composite upper bounds (46.2/44.2 at 0.8B and 59.9/55.7 at 2B in thinking/non-thinking modes), WeMM-Ecom-0.8B and WeMM-Ecom-2B maintain substantial overall leads of +6.1/+3.8 and +3.6/+4.7 points, respectively, demonstrating that domain-grounded training and distillation significantly enhance general capabilities in compact models.

\par
\ifdim\dimexpr\pagegoal-\pagetotal\relax<.5\textheight
  \pagebreak[4]
\fi
\begingroup
\captionsetup{skip=2.5pt}
\definecolor{resultgain}{HTML}{B51E32}
\definecolor{resultloss}{HTML}{23715A}
\definecolor{resulthighlight}{HTML}{EAF5EE}
\setlength{\tabcolsep}{3pt}
\setlength{\LTleft}{\fill}\setlength{\LTright}{\fill}
\setlength{\LTcapwidth}{\linewidth}
\setlength{\LTpre}{5pt}
\setlength{\LTpost}{7pt}
\renewcommand{\arraystretch}{1.00}
\newcommand{\centerresultheader}[1]{\raisebox{\dimexpr(\depth-\height)/2\relax}{#1}}
\fontsize{8.5}{10.0}\selectfont
\newcommand{\moderesult}[2]{\mbox{\fontsize{8.5}{10}\selectfont #1\,/\,#2}}
\begin{longtable}{@{}>{\raggedright\arraybackslash}p{0.235\linewidth}>{\centering\arraybackslash}p{\dimexpr0.1186\linewidth-2\tabcolsep\relax}>{\centering\arraybackslash}p{\dimexpr0.1186\linewidth-2\tabcolsep\relax}>{\centering\arraybackslash}p{\dimexpr0.1186\linewidth-2\tabcolsep\relax}>{\columncolor{resulthighlight}\centering\arraybackslash}p{\dimexpr0.1186\linewidth-2\tabcolsep\relax}>{\centering\arraybackslash}p{\dimexpr0.1186\linewidth-2\tabcolsep\relax}>{\centering\arraybackslash}p{\dimexpr0.1720\linewidth-2\tabcolsep\relax}@{}}
\caption{Comparison of dense WeMM-Ecom models with baselines on general benchmarks. Results are thinking / non-thinking. Bold and underlining mark the highest and second-highest scores in each row.}\label{tab:results_general_dense}\\
\toprule
{\bfseries Benchmark} & \multicolumn{4}{c}{\centerresultheader{{\bfseries WeMM-Ecom}}} & \centerresultheader{{\bfseries Qwen3.5}} & \centerresultheader{\mbox{\bfseries GLM-4.6V-Flash}} \\
\midrule[.35pt]
{\bfseries Size} & {\bfseries 0.8B} & {\bfseries 2B} & {\bfseries 4B} & {\bfseries 9B} & {\bfseries 9B} & {\bfseries 9B} \\
\endfirsthead
\multicolumn{7}{l}{\textbf{Table \thetable\ (continued)}} \\
\toprule
{\bfseries Benchmark} & \multicolumn{4}{c}{\centerresultheader{{\bfseries WeMM-Ecom}}} & \centerresultheader{{\bfseries Qwen3.5}} & \centerresultheader{\mbox{\bfseries GLM-4.6V-Flash}} \\
\midrule[.35pt]
{\bfseries Size} & {\bfseries 0.8B} & {\bfseries 2B} & {\bfseries 4B} & {\bfseries 9B} & {\bfseries 9B} & {\bfseries 9B} \\
\endhead
\midrule
\multicolumn{7}{r}{\footnotesize Continued on next page} \\
\endfoot
\bottomrule
\endlastfoot
\resultgroupheading{7}{STEM and Multimodal Reasoning}
CMMMU (val)~\citep{zhang2024cmmmu} & \moderesult{41.8}{34.4} & \moderesult{55.6}{51.3} & \moderesult{65.4}{59.9} & \moderesult{\underline{69.9}}{66.2} & \moderesult{{\bfseries 70.9}}{65.1} & \moderesult{61.1}{53.0} \\*
CMMU (val, MCQ)~\citep{he2024cmmu} & \moderesult{53.7}{46.3} & \moderesult{76.2}{72.4} & \moderesult{89.1}{84.3} & \moderesult{\underline{91.2}}{88.6} & \moderesult{{\bfseries 91.8}}{87.9} & \moderesult{80.4}{70.7} \\
CSVQA~\citep{jian2025csvqa} & \moderesult{25.2}{25.8} & \moderesult{51.5}{45.9} & \moderesult{71.5}{61.4} & \moderesult{{\bfseries 79.9}}{69.7} & \moderesult{\underline{77.9}}{68.8} & \moderesult{52.3}{36.6} \\
DynaMath~\citep{zou2025dynamath} & \moderesult{60.0}{49.2} & \moderesult{76.8}{71.8} & \moderesult{\underline{82.4}}{78.3} & \moderesult{{\bfseries 83.9}}{80.3} & \moderesult{81.4}{74.0} & \moderesult{73.6}{63.5} \\
LogicVista~\citep{xiao2024logicvista} & \moderesult{48.0}{39.0} & \moderesult{59.0}{57.0} & \moderesult{70.0}{67.0} & \moderesult{{\bfseries 77.0}}{71.0} & \moderesult{\underline{75.0}}{68.0} & \moderesult{60.0}{52.0} \\
MATH-Vision~\citep{wang2024measuring} & \moderesult{27.4}{27.5} & \moderesult{52.2}{50.0} & \moderesult{70.1}{65.2} & \moderesult{\underline{76.3}}{70.9} & \moderesult{{\bfseries 77.7}}{69.7} & \moderesult{55.9}{42.1} \\
MathVista (mini)~\citep{lu2024mathvista} & \moderesult{68.1}{66.1} & \moderesult{80.2}{79.0} & \moderesult{\underline{86.2}}{82.6} & \moderesult{{\bfseries 86.5}}{83.9} & \moderesult{84.6}{83.3} & \moderesult{83.0}{72.7} \\
MMMU~\citep{yue2024mmmu} & \moderesult{54.0}{42.7} & \moderesult{65.7}{63.3} & \moderesult{75.6}{72.7} & \moderesult{{\bfseries 79.1}}{76.7} & \moderesult{\underline{78.7}}{75.1} & \moderesult{72.3}{67.2} \\
MMMU-Pro~\citep{yue2025mmmu} & \moderesult{35.1}{31.5} & \moderesult{51.9}{49.8} & \moderesult{65.4}{62.4} & \moderesult{{\bfseries 69.8}}{64.9} & \moderesult{\underline{69.3}}{66.0} & \moderesult{56.9}{47.8} \\
VisualPuzzles~\citep{song2025visualpuzzles} & \moderesult{32.3}{31.3} & \moderesult{40.4}{41.7} & \moderesult{49.2}{50.2} & \moderesult{\underline{56.9}}{53.8} & \moderesult{{\bfseries 58.1}}{49.0} & \moderesult{42.0}{42.6} \\
VisuLogic~\citep{xu2026visulogic} & \moderesult{24.0}{25.7} & \moderesult{26.6}{25.8} & \moderesult{29.4}{26.7} & \moderesult{\underline{32.0}}{28.9} & \moderesult{{\bfseries 34.2}}{29.7} & \moderesult{26.6}{15.3} \\
ZeroBench (sub)~\citep{roberts2025zerobench} & \moderesult{15.3}{11.4} & \moderesult{21.9}{18.3} & \moderesult{26.7}{29.0} & \moderesult{{\bfseries 33.5}}{28.1} & \moderesult{\underline{32.3}}{29.3} & \moderesult{24.0}{16.5} \\
\resultgroupheading{7}{General VQA}
CCBench~\citep{opencompassccbench} & \moderesult{40.0}{45.3} & \moderesult{60.0}{64.7} & \moderesult{66.1}{69.4} & \moderesult{{\bfseries 74.7}}{\underline{74.5}} & \moderesult{69.6}{70.2} & \moderesult{67.8}{61.4} \\*
ChineseSimpleVQA~\citep{gu2025see} & \moderesult{17.6}{18.9} & \moderesult{32.1}{29.4} & \moderesult{45.5}{41.4} & \moderesult{\underline{49.7}}{48.5} & \moderesult{44.3}{45.7} & \moderesult{{\bfseries 50.2}}{44.6} \\
CII-Bench~\citep{zhang2025can} & \moderesult{38.4}{37.8} & \moderesult{48.9}{48.4} & \moderesult{61.3}{56.6} & \moderesult{{\bfseries 65.6}}{\underline{64.4}} & \moderesult{{\bfseries 65.6}}{63.6} & \moderesult{57.0}{54.8} \\
\mbox{MMBench\textsubscript{EN-DEV-v1.1}~\citep{liu2024mmbench}} & \moderesult{73.1}{73.0} & \moderesult{85.0}{83.0} & \moderesult{87.5}{86.7} & \moderesult{{\bfseries 90.2}}{\underline{89.6}} & \moderesult{88.2}{86.2} & \moderesult{86.2}{82.8} \\*
\mbox{MMBench\textsubscript{CN-DEV-v1.1}~\citep{liu2024mmbench}} & \moderesult{72.0}{70.3} & \moderesult{83.5}{81.6} & \moderesult{87.6}{87.3} & \moderesult{{\bfseries 89.4}}{\underline{88.9}} & \moderesult{87.2}{85.4} & \moderesult{86.3}{83.0} \\
MMStar~\citep{chen2024we} & \moderesult{62.9}{58.0} & \moderesult{72.3}{71.3} & \moderesult{76.5}{75.7} & \moderesult{{\bfseries 80.2}}{77.6} & \moderesult{\underline{77.7}}{74.6} & \moderesult{73.2}{68.1} \\
MM-Vet~\citep{yu2023mm} & \moderesult{66.0}{62.6} & \moderesult{75.6}{72.0} & \moderesult{\underline{81.5}}{76.3} & \moderesult{{\bfseries 85.9}}{80.3} & \moderesult{80.0}{76.9} & \moderesult{77.4}{76.1} \\
PerceptionBench~\citep{lin2026perceptionbench} & \moderesult{25.2}{23.6} & \moderesult{31.3}{29.0} & \moderesult{36.5}{32.8} & \moderesult{{\bfseries 41.4}}{36.5} & \moderesult{\underline{39.6}}{33.9} & \moderesult{31.2}{28.3} \\
RealWorldQA~\citep{xai2024realworldqa} & \moderesult{68.1}{66.1} & \moderesult{75.6}{71.6} & \moderesult{77.7}{78.2} & \moderesult{{\bfseries 81.1}}{77.0} & \moderesult{\underline{80.5}}{74.1} & \moderesult{75.8}{72.2} \\
SimpleVQA~\citep{cheng2025simplevqa} & \moderesult{42.7}{37.9} & \moderesult{54.8}{48.3} & \moderesult{\underline{59.1}}{53.7} & \moderesult{{\bfseries 59.9}}{57.6} & \moderesult{56.4}{53.8} & \moderesult{58.4}{53.3} \\
WorldVQA~\citep{zhou2026worldvqa} & \moderesult{8.2}{6.5} & \moderesult{14.1}{13.8} & \moderesult{20.6}{16.7} & \moderesult{{\bfseries 22.1}}{20.1} & \moderesult{\underline{20.7}}{19.4} & \moderesult{14.9}{12.1} \\
\resultgroupheading{7}{Documents and Charts}
AI2D (test)~\citep{kembhavi2016diagram} & \moderesult{73.9}{73.1} & \moderesult{83.7}{82.7} & \moderesult{87.8}{86.2} & \moderesult{{\bfseries 89.4}}{\underline{88.3}} & \moderesult{{\bfseries 89.4}}{86.8} & \moderesult{84.7}{84.8} \\*
CC-OCR~\citep{yang2025cc} & \moderesult{73.6}{77.6} & \moderesult{80.6}{81.8} & \moderesult{82.4}{84.6} & \moderesult{84.7}{{\bfseries 85.6}} & \moderesult{84.3}{\underline{85.3}} & \moderesult{63.7}{62.8} \\*
ChartQAPro~\citep{masry2025chartqapro} & \moderesult{49.0}{33.8} & \moderesult{58.4}{56.2} & \moderesult{65.6}{64.7} & \moderesult{{\bfseries 68.2}}{66.8} & \moderesult{\underline{67.0}}{64.3} & \moderesult{58.1}{43.4} \\*
ChartQA (test)~\citep{masry2022chartqa} & \moderesult{87.0}{82.6} & \moderesult{89.7}{87.1} & \moderesult{\underline{90.6}}{87.2} & \moderesult{{\bfseries 90.8}}{87.8} & \moderesult{89.9}{86.6} & \moderesult{89.0}{88.2} \\*
CharXiv (RQ)~\citep{wang2024charxiv} & \moderesult{50.7}{45.3} & \moderesult{60.9}{59.5} & \moderesult{70.3}{64.4} & \moderesult{\underline{73.1}}{68.6} & \moderesult{{\bfseries 74.3}}{67.9} & \moderesult{60.7}{50.3} \\*
DocVQA (val)~\citep{mathew2021docvqa} & \moderesult{94.1}{93.8} & \moderesult{96.2}{96.4} & \moderesult{97.1}{96.6} & \moderesult{{\bfseries 97.5}}{97.0} & \moderesult{\underline{97.3}}{96.1} & \moderesult{96.5}{96.5} \\*
InfographicVQA (val)~\citep{mathew2022infographicvqa} & \moderesult{77.8}{72.8} & \moderesult{87.7}{84.6} & \moderesult{90.8}{87.9} & \moderesult{{\bfseries 92.2}}{90.9} & \moderesult{\underline{91.8}}{88.3} & \moderesult{89.0}{86.2} \\*
OCRBench~\citep{liu2024ocrbench} & \moderesult{80.6}{84.6} & \moderesult{88.0}{89.0} & \moderesult{86.2}{86.7} & \moderesult{\underline{89.8}}{{\bfseries 90.6}} & \moderesult{87.8}{85.2} & \moderesult{85.4}{85.3} \\*
OCRBench v2~\citep{fu2026ocrbench} & \moderesult{50.1}{48.2} & \moderesult{56.3}{51.3} & \moderesult{60.8}{57.9} & \moderesult{{\bfseries 65.6}}{60.9} & \moderesult{\underline{64.0}}{62.1} & \moderesult{53.2}{50.3} \\*
OmniDocBench~\citep{ouyang2025omnidocbench} & \moderesult{80.1}{84.3} & \moderesult{85.9}{86.9} & \moderesult{86.1}{\underline{90.7}} & \moderesult{{\bfseries 90.9}}{90.5} & \moderesult{86.8}{88.6} & \moderesult{53.5}{50.6} \\*
WildDoc~\citep{wang2025wilddoc} & \moderesult{61.3}{54.7} & \moderesult{67.7}{52.8} & \moderesult{67.7}{65.5} & \moderesult{\underline{70.4}}{69.2} & \moderesult{{\bfseries 71.8}}{\underline{70.4}} & \moderesult{70.0}{64.6} \\
\resultgroupheading{7}{Grounding and Counting}
CountBenchQA~\citep{beyer2024paligemma} & \moderesult{75.8}{72.1} & \moderesult{90.0}{85.1} & \moderesult{93.3}{84.5} & \moderesult{\underline{95.5}}{90.8} & \moderesult{{\bfseries 97.0}}{87.8} & \moderesult{90.6}{83.1} \\*
CountQA~\citep{tamarapalli2025countqa} & \moderesult{18.3}{12.4} & \moderesult{37.0}{31.2} & \moderesult{46.8}{33.3} & \moderesult{{\bfseries 65.3}}{38.6} & \moderesult{\underline{61.8}}{37.4} & \moderesult{35.9}{27.6} \\
MIHBench~\citep{li2025mihbench} & \moderesult{78.6}{42.6} & \moderesult{\underline{83.1}}{53.5} & \moderesult{{\bfseries 83.6}}{52.4} & \moderesult{82.6}{56.9} & \moderesult{82.6}{51.9} & \moderesult{81.1}{80.6} \\
ODInW13~\citep{li2022grounded} & \moderesult{33.0}{26.4} & \moderesult{40.5}{37.9} & \moderesult{41.8}{40.6} & \moderesult{{\bfseries 44.5}}{\underline{43.2}} & \moderesult{41.8}{42.8} & \moderesult{26.3}{26.2} \\
PixMo-Count~\citep{deitke2025molmo} & \moderesult{52.9}{42.0} & \moderesult{75.6}{70.0} & \moderesult{80.9}{63.7} & \moderesult{{\bfseries 87.8}}{68.3} & \moderesult{\underline{87.6}}{74.2} & \moderesult{73.9}{57.6} \\
RefCOCO (avg)~\citep{yu2016modeling,mao2016generation} & \moderesult{76.8}{78.8} & \moderesult{82.8}{81.9} & \moderesult{85.2}{84.8} & \moderesult{\underline{86.4}}{86.3} & \moderesult{{\bfseries 86.9}}{86.2} & \moderesult{84.1}{83.8} \\
Ref-L4 (val)~\citep{chen2025revisiting} & \moderesult{78.6}{83.6} & \moderesult{85.8}{87.5} & \moderesult{86.0}{89.0} & \moderesult{\underline{89.5}}{{\bfseries 90.0}} & \moderesult{88.1}{88.2} & \moderesult{88.2}{88.2} \\
TallyQA~\citep{acharya2019tallyqa} & \moderesult{75.4}{75.0} & \moderesult{80.3}{79.9} & \moderesult{81.4}{81.5} & \moderesult{{\bfseries 82.8}}{\underline{82.1}} & \moderesult{79.7}{80.2} & \moderesult{81.8}{81.1} \\
\resultgroupheading{7}{2D/3D Spatial Understanding}
CV-Bench~\citep{tong2024cambrian} & \moderesult{80.3}{76.4} & \moderesult{86.1}{86.1} & \moderesult{88.0}{86.3} & \moderesult{{\bfseries 88.8}}{87.0} & \moderesult{\underline{88.3}}{86.1} & \moderesult{86.1}{83.5} \\*
EmbSpatial-Bench~\citep{du2024embspatial} & \moderesult{72.1}{63.8} & \moderesult{79.5}{75.7} & \moderesult{81.3}{76.1} & \moderesult{{\bfseries 82.2}}{76.2} & \moderesult{\underline{81.7}}{72.6} & \moderesult{80.8}{78.2} \\
ERQA~\citep{team2025gemini} & \moderesult{40.5}{34.3} & \moderesult{44.8}{43.8} & \moderesult{51.0}{45.3} & \moderesult{\underline{53.3}}{43.5} & \moderesult{{\bfseries 55.3}}{45.5} & \moderesult{43.8}{41.3} \\
MMSI-Bench~\citep{yang2026mmsi} & \moderesult{27.8}{29.2} & \moderesult{30.2}{28.3} & \moderesult{33.8}{33.0} & \moderesult{{\bfseries 38.1}}{32.6} & \moderesult{\underline{36.3}}{31.9} & \moderesult{30.3}{31.3} \\
OmniSpatial~\citep{jia2026omnispatial} & \moderesult{43.3}{40.8} & \moderesult{49.1}{51.0} & \moderesult{54.1}{50.9} & \moderesult{\underline{56.2}}{53.2} & \moderesult{{\bfseries 56.5}}{48.0} & \moderesult{48.8}{45.7} \\
SpatialMQA~\citep{liu2025can} & \moderesult{40.7}{37.6} & \moderesult{45.0}{43.6} & \moderesult{51.5}{47.3} & \moderesult{{\bfseries 57.1}}{45.0} & \moderesult{\underline{57.0}}{45.1} & \moderesult{46.2}{42.3} \\
\resultgroupheading{7}{Multi-Image Understanding}
BLINK~\citep{fu2024blink} & \moderesult{57.9}{52.8} & \moderesult{64.2}{63.8} & \moderesult{66.1}{65.8} & \moderesult{\underline{68.5}}{68.1} & \moderesult{{\bfseries 68.9}}{65.2} & \moderesult{66.0}{64.0} \\*
Mantis-Eval~\citep{jiang2024mantis} & \moderesult{76.0}{57.1} & \moderesult{82.5}{81.6} & \moderesult{87.1}{83.0} & \moderesult{{\bfseries 88.0}}{81.6} & \moderesult{\underline{87.6}}{82.0} & \moderesult{83.4}{77.4} \\
MMIU~\citep{meng2025mmiu} & \moderesult{54.1}{46.0} & \moderesult{61.3}{58.5} & \moderesult{65.8}{63.6} & \moderesult{{\bfseries 68.8}}{64.5} & \moderesult{\underline{67.0}}{59.8} & \moderesult{62.4}{59.4} \\
MuirBench~\citep{wang2025muirbench} & \moderesult{52.1}{35.1} & \moderesult{59.6}{59.8} & \moderesult{65.5}{66.2} & \moderesult{\underline{71.4}}{63.0} & \moderesult{67.3}{52.4} & \moderesult{{\bfseries 75.6}}{63.5} \\
\resultgroupheading{7}{Instruction Following}
CFBench~\citep{zhang2025cfbench} & \moderesult{64.6}{57.4} & \moderesult{78.0}{70.0} & \moderesult{\underline{84.5}}{80.5} & \moderesult{\underline{84.5}}{81.3} & \moderesult{{\bfseries 84.7}}{83.0} & \moderesult{81.8}{74.5} \\*
IFBench~\citep{pyatkin2026generalizing} & \moderesult{29.7}{28.0} & \moderesult{45.3}{33.3} & \moderesult{56.3}{32.7} & \moderesult{\underline{68.0}}{50.7} & \moderesult{{\bfseries 69.3}}{40.0} & \moderesult{28.3}{22.0} \\
IFEval~\citep{zhou2023instruction} & \moderesult{69.9}{60.8} & \moderesult{84.8}{73.9} & \moderesult{89.3}{79.9} & \moderesult{\underline{91.5}}{82.3} & \moderesult{{\bfseries 92.4}}{84.5} & \moderesult{63.4}{57.3} \\
MIA-Bench~\citep{qian2025mia} & \moderesult{67.5}{68.9} & \moderesult{80.6}{76.5} & \moderesult{\underline{83.5}}{81.0} & \moderesult{83.3}{{\bfseries 83.7}} & \moderesult{83.1}{82.7} & \moderesult{81.1}{77.0} \\
MM-IFEval~\citep{ding2025mm} & \moderesult{56.0}{48.2} & \moderesult{67.3}{58.5} & \moderesult{77.9}{68.2} & \moderesult{\underline{81.1}}{70.5} & \moderesult{{\bfseries 81.3}}{64.8} & \moderesult{66.1}{57.7} \\
\resultgroupheading{7}{Text-Centric Tasks}
C-Eval~\citep{huang2023c} & \moderesult{58.3}{52.5} & \moderesult{74.8}{69.9} & \moderesult{85.8}{81.9} & \moderesult{{\bfseries 88.2}}{86.3} & \moderesult{\underline{88.0}}{86.7} & \moderesult{84.7}{77.6} \\*
GPQA (Diamond)~\citep{rein2023gpqa} & \moderesult{35.4}{36.4} & \moderesult{60.1}{54.6} & \moderesult{71.7}{67.2} & \moderesult{75.8}{75.8} & \moderesult{\underline{78.3}}{{\bfseries 78.8}} & \moderesult{41.4}{42.4} \\
HMMT (Nov.\ 2025)~\citep{dekoninck2026beyond} & \moderesult{10.0}{3.3} & \moderesult{27.5}{33.3} & \moderesult{70.8}{68.3} & \moderesult{\underline{80.0}}{75.8} & \moderesult{{\bfseries 80.8}}{61.7} & \moderesult{19.2}{9.2} \\
IMO-AnswerBench (v2)~\citep{luong2025towards} & \moderesult{8.5}{6.8} & \moderesult{31.5}{32.5} & \moderesult{59.3}{55.5} & \moderesult{\underline{66.3}}{63.3} & \moderesult{{\bfseries 66.8}}{48.3} & \moderesult{25.8}{20.8} \\
MMLU-Pro~\citep{wang2024mmlu} & \moderesult{49.6}{47.6} & \moderesult{69.2}{66.2} & \moderesult{80.0}{77.6} & \moderesult{{\bfseries 82.8}}{81.1} & \moderesult{\underline{82.5}}{80.4} & \moderesult{73.6}{65.4} \\
MMLU-Redux~\citep{gema2025we} & \moderesult{61.5}{53.4} & \moderesult{80.5}{76.1} & \moderesult{89.4}{85.9} & \moderesult{{\bfseries 92.0}}{89.7} & \moderesult{\underline{91.5}}{90.1} & \moderesult{87.0}{69.9} \\
SuperGPQA~\citep{du2026supergpqa} & \moderesult{25.9}{23.8} & \moderesult{40.8}{36.7} & \moderesult{53.7}{49.5} & \moderesult{{\bfseries 58.7}}{55.2} & \moderesult{\underline{58.6}}{55.3} & \moderesult{45.9}{34.8} \\
\midrule
{\bfseries Overall mean (64)} & \moderesult{52.3}{48.0} & \moderesult{63.5}{60.4} & \moderesult{70.5}{66.6} & \moderesult{{\bfseries 74.2}}{69.7} & \moderesult{\underline{73.4}}{67.6} & \moderesult{63.3}{57.9} \\
\end{longtable}
\endgroup

\section{Conclusion}
We presented \model, a family of five multimodal large language models ranging from a dense 0.8B model to a 35B-A3B mixture-of-experts model, together with a 28-benchmark e-commerce evaluation suite that spans eight business scenarios and 207,117 instances. Each model is trained with two stages of continued pre-training that raise the e-commerce share from 45.8\% to 61.1\%, and then post-trained with hybrid-thinking on-policy distillation, a single parameter-space merge with the post-trained Qwen3.5 backbone, and GRPO, yielding one checkpoint that serves both thinking and non-thinking modes. Across all ten size--mode comparisons, \model\ improves on its Qwen3.5 baseline matched to size in e-commerce while matching or surpassing it on average across the 64 general benchmarks. We also show that WeMM-Ecom-35B-A3B attains the highest e-commerce average among evaluated open-source flagships with $5\times$--$22\times$ fewer total parameters, and WeMM-Ecom-9B surpasses the $3\times$ larger Qwen3.8-27B. This indicates that domain specialization does not necessarily come at the expense of general capabilities, and scale does not gate the specialization benefit. 

We release all model weights together with our data curation, training recipe, and benchmark construction so that this recipe can be examined and extended in future endeavors regarding domain-specialized models. In production, \model\ already serves hundreds of workflows in WeChat E-Commerce, handling billion-scale requests per day across catalog management, multimodal product search, and livestream commerce. Future work will extend \model\ to interactive shopping agents and study hardware-aware quantization for real-time edge deployment.

\FloatBarrier

\begingroup
\setlength{\bibsep}{2pt plus 0.5pt minus 0.5pt}
\appto\bibfont{%
  \clubpenalties 2 10000 0\relax
  \widowpenalties 2 10000 0\relax
}
\bibliographystyle{plainnat}
\bibliography{main}
\endgroup

\clearpage
\appendix
\section{Training and Data Details}
\label{app:repro}

\subsection{Post-Training Hyperparameters and Prompt Curation}
\label{app:post_training_settings}

\paragraph{Distillation and RL sampling.}
\label{app:compact_rl}
During OPD, we sample two responses per prompt at temperature 1.0 and top-$p$ 1.0, with a maximum generation length of 24,576 tokens within a 32,768-token sequence limit. The distillation loss weight is set to 1.0 without an additional reference-policy penalty; compact students (0.8B, 2B, and 4B) initialize from their CPT checkpoints with zero weight decay. During RL, we sample $G=16$ responses per prompt within a 65,536-token context limit using weight decay 0.01, temperature 1.0, and top-$p$ 0.9 by default (supplemented with top-$k$ 20 and repetition penalty 1.05 on instruction-following tasks). All RL runs initialize both the actor and frozen reference policy from the equally weighted ($\alpha=0.5$) post-OPD merged checkpoint.

\paragraph{Post-training prompt screening.}
\label{app:post_training_screening}
To construct the OPD prompt pool, both the teacher and student models generate eight candidate responses per prompt; we retain prompts where the teacher achieves a strictly higher mean quality score than the student. For the RL pool, the corresponding post-OPD merged model generates eight thinking-mode rollouts per candidate prompt. We retain prompts with positive reward variance and a mean task reward of at most 0.95, and additionally require a strictly positive mean reward on mathematical and logical reasoning tasks.

\paragraph{Mode assignment.}
\label{app:mode_assignment}
Operating modes are assigned deterministically by hashing each prompt identifier so that all rollouts for a given prompt share the same mode. Thinking mode accounts for 400,229 prompts (80.05\%) in the 499,998-prompt OPD pool and 24,823 prompts (80.14\%) in the 30,976-prompt RL pool.

\subsection{Reward Design and RL Optimization}
\label{app:rl_details}
In Equation~\ref{eq:adv}, group standard deviations are computed using Bessel's correction with stabilizer $\epsilon_A=10^{-4}$ (groups with identical rewards receive zero advantage). In Equation~\ref{eq:rl}, the token-level KL penalty is computed via the Schulman approximator $K_{i,t}=\min\!\big(10,\exp(d_{i,t})-d_{i,t}-1\big)$ with $d_{i,t}=\operatorname{clip}\!\big(\log\pi_{\mathrm{ref}}(y_{i,t}\mid h_{i,t})-\ell_{i,t},-20,20\big)$ and KL coefficient 0.01.

\paragraph{E-commerce and general task rewards.}
\label{app:commerce_rewards}
\label{app:if_reward}
Task rewards combine deterministic rule verifiers with model-based judgments from Qwen3.5-122B-A10B. Product and SKU retrieval tasks assign unit reward for predicting the correct equivalence label in the required schema. For localization and detection tasks, let $N$ denote the total number of predicted and ground-truth boxes and $\mathcal P$ denote the optimal one-to-one matching within each image. For $N>0$, the detection reward is
\begin{equation}
 R_{\mathrm{det}}=\frac{2}{N}\sum_{(p,g)\in\mathcal P}
 \operatorname{IoU}(p,g)\,w(p,g),
 \label{eq:det_reward}
\end{equation}
where $w(p,g)=1$ for livestream and multi-image detection, and $w(p,g)=0.7+0.3\,s(p,g)$ for target-product detection ($s=1$ for exact class match, $0.5$ for partial/full product class confusion, and $0$ otherwise). For instruction-following tasks, full constraint satisfaction receives reward 1.0, while partial compliance receives $0.49$ times the mean constraint satisfaction score evaluated on the visible final answer after stripping the reasoning block.

\paragraph{Format validation.}
Rollouts that violate the required output schema or hit the generation cap before completion receive zero task reward; in thinking mode, responses must contain a well-formed reasoning block followed by a non-empty final answer. If verifier evaluation fails on a prompt group, the task-reward term is masked out and only the reference KL penalty is applied.

\section{Evaluation Benchmarks and Protocols}
\label{app:eval_details}
\label{app:recorded_metrics}
\label{sec:scoring}

\paragraph{Decoding and semantic adjudication.}
For \model, thinking mode uses temperature 1.0, top-$p$ 0.95, and top-$k$ 20, paired with presence penalty 2.0 and repetition penalty 1.05 on reasoning-intensive benchmarks and presence penalty 1.5 with repetition penalty 1.0 on the remaining tasks. In non-thinking mode, standard benchmarks use temperature 0.7 and top-$p$ 0.8, while open-ended generation subsets use temperature 1.0, top-$p$ 1.0, and top-$k$ 40. All models use a default maximum generation limit of 65,536 tokens (with greedy decoding at 32,768 tokens for OmniDocBench). Baseline models are evaluated under their recommended or task-calibrated decoding parameters. Model-based evaluation adjudication uses Qwen3-30B-A3B-Instruct-2507~\citep{yang2025qwen3}.

\begingroup
\setlength{\tabcolsep}{3pt}
\renewcommand{\arraystretch}{1.06}
\small
\setlength{\LTleft}{\fill}
\setlength{\LTright}{\fill}
\begin{longtable}{@{}>{\raggedright\arraybackslash}p{0.235\linewidth}>{\raggedleft\arraybackslash}p{0.09\linewidth}>{\raggedright\arraybackslash}p{0.423\linewidth}>{\raggedright\arraybackslash}p{0.195\linewidth}@{}}
\caption{E-commerce evaluation sets and scoring metrics.}
\label{tab:commerce_protocols}\\
\toprule
Benchmark & Instances & Input and target & Primary metric \\
\midrule
\endfirsthead
\multicolumn{4}{l}{\textbf{Table \thetable\ (continued)}}\\
\toprule
Benchmark & Instances & Input and target & Primary metric \\
\midrule
\endhead
\midrule
\multicolumn{4}{r}{\footnotesize Continued on next page}\\
\endfoot
\bottomrule
\multicolumn{4}{@{}p{\dimexpr0.943\linewidth+6\tabcolsep\relax}@{}}{%
  \begin{papertablenotes}
    \item Each instance may contain several images, targets, or products.
  \end{papertablenotes}%
} \\
\endlastfoot
\multicolumn{4}{l}{\textbf{Product Knowledge}}\\*
E-commerce Knowledge & 4,048 & Text questions about products, brands, and industry knowledge & Judged accuracy \\
Entity Recognition & 550 & Product image to product name, brand, and model & Acceptability \\
\addlinespace[2pt]
\multicolumn{4}{l}{\textbf{Product Attributes}}\\*
Attribute Selection & 3,020 & Product image and candidate values to one attribute value & Accuracy \\
Attribute Extraction & 14,534 & Listing images, product fields, and options to attribute values & Reference-attribute accuracy \\
Brand Attribution & 6,122 & Listing images and fields, with optional brand candidates, to product brand & Accuracy \\
Brand Extraction & 3,000 & Listing images and product context to an extracted brand name & Accuracy \\
\addlinespace[2pt]
\multicolumn{4}{l}{\textbf{Product Moderation}}\\*
Illegal Goods & 3,496 & Listing evidence and rules to an illegal-goods decision & Accuracy \\
Sales Restrictions & 1,050 & Listing evidence and rules to a restricted-sales decision & Accuracy \\
Image-Title Mismatch & 1,897 & Product title and images to the set of unrelated image indices & Exact-set accuracy \\
\addlinespace[2pt]
\multicolumn{4}{l}{\textbf{Product OCR}}\\*
Product OCR & 19,962 & Single, multiple, or long product images to an answer about visible text & Mean judged accuracy \\
\addlinespace[2pt]
\multicolumn{4}{l}{\textbf{Product Retrieval}}\\*
Region-level Retrieval & 15,335 & Two product crops to an ordinal similarity judgment & Normalized Spearman \\
Product-level Retrieval & 2,981 & Two product images and listing fields to a same-product judgment & Average precision \\
SKU-level Retrieval & 3,268 & Two textual SKU descriptions to a variant-equivalence decision & Match F1 \\
\addlinespace[2pt]
\multicolumn{4}{l}{\textbf{Product Grounding}}\\*
Chart Detection & 1,456 & Listing image to chart and table boxes & Mean class F1 \\
Product Detection & 5,000 & Product image to boxes for the target product & Box F1 \\
Logo Detection & 1,500 & Product image to brand-logo boxes & Box F1 \\
Qualification Detection & 997 & Product image to typed certification-mark boxes & Mean class F1 \\
Dimension Detection & 994 & Product image to dimension-annotation boxes & Box F1 \\
Livestream Detection & 1,992 & Livestream frame to boxes for the advertised product & Box F1 \\
\addlinespace[2pt]
\multicolumn{4}{l}{\textbf{E-commerce Reasoning}}\\*
Chart Reasoning & 2,000 & Rendered SKU, price, attribute, or constraint table to a derived answer & Mean accuracy \\
Category Reasoning & 500 & Product image and options to the requested category & Accuracy \\
Product Counting & 500 & Product arrangement and selection constraints to an item count & Accuracy \\
Spatial Reasoning & 500 & Product arrangement and relative-position constraints to a target item & Accuracy \\
\addlinespace[2pt]
\multicolumn{4}{l}{\textbf{E-commerce (Open)}}\\*
ShoppingMMLU & 20,570 & Text tasks spanning shopping knowledge and e-commerce operations & Mean task score \\
RAIR & 48,835 & Query, item fields, and available images to a relevance grade & Macro F1 \\
ChineseEcomQA & 1,800 & Chinese e-commerce questions and task instructions to answers or labels & Mean task score \\
EcomMMMU (GTS) & 36,736 & Main product image and text to outputs for eight GTS tasks & GTS mean score \\
IndustryBench-MIPU & 4,474 & Multiple industrial-product images and a request to attribute values & Micro F1 \\
\end{longtable}
\endgroup

\paragraph{Benchmark construction and answer adjudication.}
Table~\ref{tab:commerce_protocols} summarizes the 28 e-commerce benchmarks. Product OCR averages accuracies equally across single-image (4,970 instances), multi-image (6,494), and long-image (8,498) subsets. Chart Reasoning averages four 500-question subsets covering SKU charts, price tables, attribute tables, and constraint tables; questions are synthesized under category-specific templates with programmatically computed reference answers and cross-checked by frontier LLMs for reasoning consistency. During evaluation, rule-based matchers first check typed answers and taxonomy aliases before invoking the LLM judge for semantic equivalence. Entity Recognition credits full or partial identifications, Brand tasks allow semantic alias matching, and Image-Title Mismatch requires exact set agreement.

\paragraph{Grounding evaluation protocol.}
\label{app:grounding_protocol}
All six grounding benchmarks use an IoU threshold of 0.5 with coordinates normalized to $[0,1000]$. Chart Detection and Qualification Detection report macro-averaged class F1, whereas the remaining sets report box F1 (with Product Detection using greedy one-to-one matching). In Chart Detection, prompts request JSON objects with \texttt{bbox\_2d} and \texttt{category}; to ensure schema compatibility across model families, the evaluation parser also accepts class labels output under the \texttt{label} key when \texttt{category} is absent.

\paragraph{Retrieval and public benchmark protocols.}
In Region-level Retrieval, Spearman rank correlation $\rho$ is linearly rescaled to $50(\rho+1)$; for Region-level and Product-level Retrieval, non-thinking mode scores token probabilities whereas thinking mode parses generated labels. Among the public benchmarks, EcomMMMU~\citep{ling2025ecommmmu} is evaluated on its labeled general test set (GTS) using the main product image and reported as the unweighted mean across its eight subtasks, while IndustryBench-MIPU~\citep{qi2026industrybench} uses multi-image product inputs and reports attribute-extraction micro F1.

\end{document}